\documentclass[10pt,twocolumn,letterpaper]{article}

\usepackage[pagenumbers]{cvpr}
\usepackage{times}
\usepackage{epsfig}
\usepackage{graphicx}
\usepackage{amsmath}
\usepackage{amssymb}
\usepackage{comment}
\usepackage{booktabs} %
\usepackage{numprint}
\usepackage{balance}
\usepackage[accsupp]{axessibility}  %

\DeclareMathOperator*{\argmin}{arg\,min}

\usepackage[pagebackref=true,breaklinks=true,letterpaper=true,colorlinks,bookmarks=false]{hyperref}

\def\cvprPaperID{4811} %
\def\confName{CVPR}
\def\confYear{2022}

\newcommand{\qheading}[1]{\noindent\textbf{#1}}
\newcommand{\model}{EMOCA\xspace}
\newcommand{\modellong}{\model~(EMOtion Capture and Animation)}

\begin{document}

\title{EMOCA: Emotion Driven Monocular Face Capture and Animation}

\author{Radek Daněček\\
{\tt\small rdanecek@tue.mpg.de}
\and
Michael J. Black\\
{\tt\small black@tue.mpg.de}
\and
Timo Bolkart\\
{\tt\small tbolkart@tue.mpg.de}
\and
Max Planck Institute for Intelligent Systems, %
T{\"u}bingen, Germany
}
\twocolumn[{%
\maketitle

\newcommand{\teaserwidth}{0.085}

\renewcommand\twocolumn[1][]{#1}
\begin{center}
\vspace{-0.19in}
    \centering
    \captionsetup{type=figure}
    \includegraphics[width=\teaserwidth\textwidth]{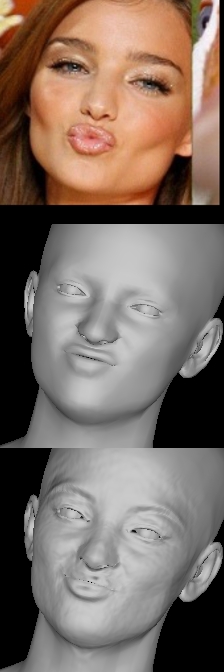}
    \includegraphics[width=\teaserwidth\textwidth]{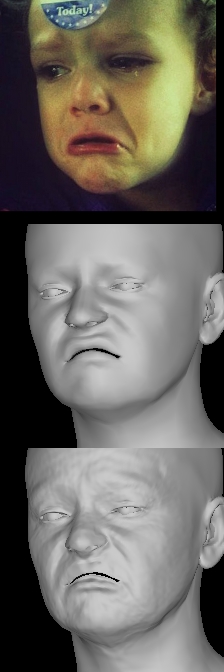}
    \includegraphics[width=\teaserwidth\textwidth]{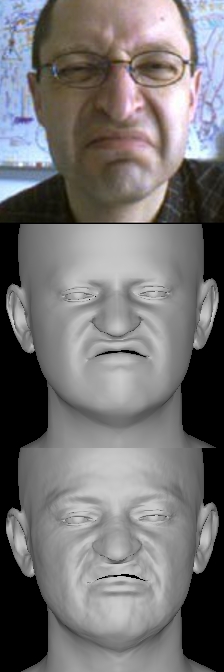}
    \includegraphics[width=\teaserwidth\textwidth]{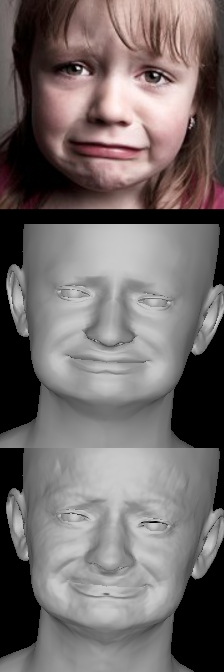}
    \includegraphics[width=\teaserwidth\textwidth]{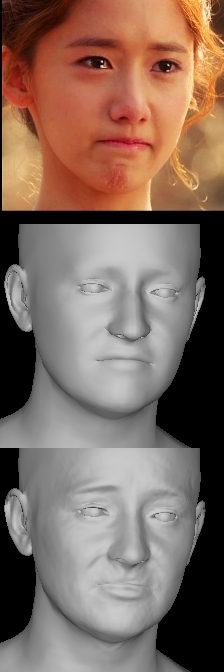}
    \includegraphics[width=\teaserwidth\textwidth]{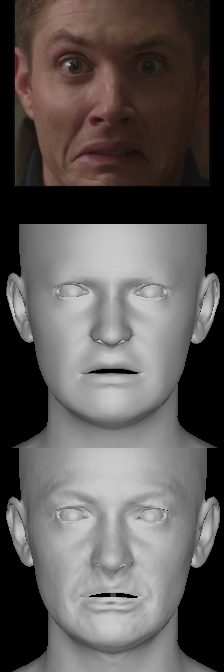}
    \includegraphics[width=\teaserwidth\textwidth]{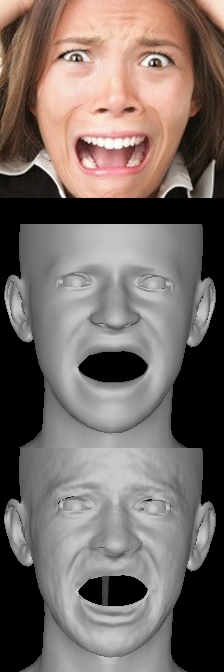}
    \includegraphics[width=\teaserwidth\textwidth]{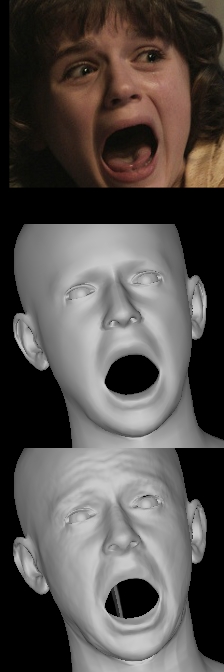}
    \includegraphics[width=\teaserwidth\textwidth]{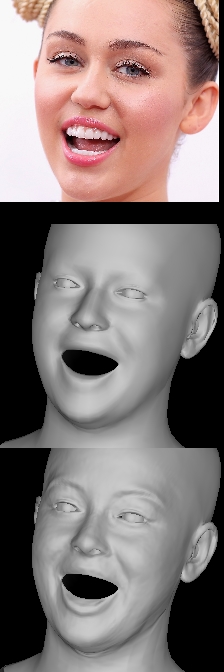}
    \includegraphics[width=\teaserwidth\textwidth]{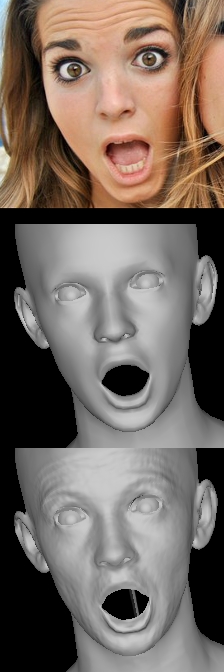}
    \includegraphics[width=\teaserwidth\textwidth]{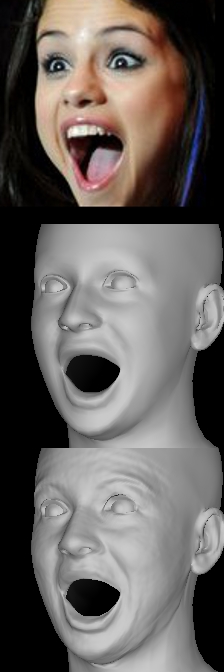}
	\vspace{-0.05in}
	\captionof{figure}{\textbf{\model} regresses 3D faces from images with facial geometry that captures the original emotional content.  Top row: images of people with challenging expressions. Middle row: coarse shape reconstruction. Bottom row: reconstruction with detailed displacements.}
    \label{fig:teaser}
\end{center}
}]

\begin{abstract}
\vspace*{-0.69em}
As 3D facial avatars become more widely used for communication, it is critical that they faithfully convey emotion.
Unfortunately, the best recent methods that regress parametric 3D face models from monocular images are unable to capture the full spectrum of facial expression, such as subtle or extreme emotions.
We find the standard reconstruction metrics used for training (landmark reprojection error, photometric error, and face recognition loss) are insufficient to capture high-fidelity expressions. 
The result is facial geometries that do not match the emotional content of the input image.
We address this with \modellong, by introducing a novel deep perceptual emotion consistency loss during training, which helps ensure that the reconstructed 3D expression matches the expression depicted in the input image. 
While \model~achieves 3D reconstruction errors that are on par with the current best methods, it significantly outperforms them in terms of the quality of the reconstructed expression and the perceived emotional content.
We also directly regress levels of valence and arousal and classify basic expressions from the estimated 3D face parameters.
On the task of in-the-wild emotion recognition, our purely geometric approach is on par with the best image-based methods, 
highlighting the value of 3D geometry in analyzing human behavior.
The model and code are publicly available at \url{https://emoca.is.tue.mpg.de}.

\end{abstract}

\newcommand{\vect}[1]{\mathbf{#1}}

\newcommand{\norm}[1]{\left\lVert#1\right\rVert}

\newcommand{\shapecoeff}{\boldsymbol{\beta}}
\newcommand{\shapedim}{{\left| \shapecoeff \right|}}
\newcommand{\shapespace}{\mathcal{S}}
\newcommand{\shapespaceexpl}{\mathbb{R}^{\shapedim}}
\newcommand{\posecoeff}{\boldsymbol{\theta}}
\newcommand{\posedim}{{\left| \posecoeff \right|}}
\newcommand{\posespace}{\mathcal{P}}
\newcommand{\posespaceexpl}{\mathbb{R}^{\posedim}}
\newcommand{\expcoeff}{\boldsymbol{\psi}}
\newcommand{\expdim}{{\left| \expcoeff \right|}}
\newcommand{\expspace}{\mathcal{E}}
\newcommand{\expspaceexpl}{\mathbb{R}^{\expdim}}
\newcommand{\numverts}{n_v}
\newcommand{\numfaces}{n_f}
\newcommand{\template}{\textbf{T}}

\newcommand{\numjoints}{k}
\newcommand{\joints}{\textbf{J}}
\newcommand{\jointregressor}{\mathcal{J}}
\newcommand{\blendweights}{\mathcal{W}}
\newcommand{\blendweightsdim}{\left| \mathcal{W} \right|}

\newcommand{\landmark}{\textbf{k}}

\newcommand{\lighting}{\textbf{l}}
\newcommand{\cam}{\textbf{c}}

\newcommand{\albedo}{A}
\newcommand{\albedocoeffs}{\boldsymbol{\alpha}}
\newcommand{\albedodim}{\left| \albedocoeffs \right|}
\newcommand{\albedospaceexpl}{\mathbb{R}^{\albedodim}}
\newcommand{\normalcoeffs}{\boldsymbol{\nu}}
\newcommand{\normaldim}{\left| \normalcoeffs \right|}

\newcommand{\uvsize}{d}
\newcommand{\image}{I}

\newcommand{\flamev}{M}
\newcommand{\normals}{N}

\newcommand{\displacements}{D}
\newcommand{\detailgeom}{G}

\newcommand{\coarseencoder}{E_c}
\newcommand{\coarseencoderexp}{E_e}
\newcommand{\detailencoder}{E_d}
\newcommand{\edm}{F_d}

\newcommand{\zcode}{\boldsymbol{\delta}}
\newcommand{\zcodedim}{\left| \zcode \right|}
\newcommand{\zcodeexpl}{\mathbb{R}^{\zcodedim}}
\newcommand{\cons}{constraint}

\newcommand{\coarsedecoder}{R_c}
\newcommand{\renderer}{R}

\newcommand{\emonet}{A}
\newcommand{\emovec}{\boldsymbol{\epsilon}}
\newcommand{\emometric}{d}

\newcommand{\secref}[1]{Sec.~\ref{#1}}
\newcommand{\figref}[1]{Fig.~\ref{#1}}
\newcommand{\tabref}[1]{Tab.~\ref{#1}}
\newcommand{\supmat}{Sup.~Mat.\xspace}
\section{Introduction}

Teaching computers to see humans and understand their behavior is a long-standing goal of computer vision. 
To accomplish this, computers need to understand how humans look, how they move, and what they feel. 
Faces and their emotional expressions provide an important source of information about a person's internal emotional state. 
To support automated analysis of emotional state, we capture a person's face, including its 3D shape, pose, and facial expression, given a single RGB image. 
To do so, we go beyond prior work to extract 3D geometry that carries rich emotional content.
We focus on parametric methods (i.e., animatable and model-based) due to their wide applicability for 3D avatar creation \cite{Hu2017}, image synthesis \cite{Ghosh2020,Tewari2020}, video editing \cite{Thies2016,Kim2018_DeepVideo} and face recognition \cite{Blanz2002,Romdhani2002}. 

The field of 3D face reconstruction has rapidly advanced over the last two decades; see Egger et al.~\cite{Egger2020} for a review.
Existing methods that estimate 3D face models struggle to capture facial expressions in detail and often produce 3D shapes that do not carry the emotional content of the input image.
This has several causes. 
First, some 3D face models  lack sufficient expressiveness to capture subtle or extreme expressions.
Second, reconstruction metrics like landmark reprojection loss \cite{Blanz2003}, photometric loss \cite{BlanzVetter1999}, face recognition loss \cite{Genova2018}, or multi-image consistency losses \cite{Sanyal2019_RingNet,Tewari2019_FML}, are either not affected by facial expressions, or require perfect image alignment to capture subtle cues. 
Subtle changes in geometry, however, can result in large differences in the perceived emotion.
We argue that, to recover 3D expression accurately, 
we need a new reconstruction metric that measures differences in expressions between the 3D reconstruction and the input image.

To that end, we describe \modellong, a neural network that learns an animatable %
face model from in-the-wild images without 3D supervision. 
The design of our method is inspired by advances in the field of facial emotion recognition, 
which has made tremendous progress to date on estimating affect (or emotion) from in-the-wild-images \cite{Li2020}.
Specifically, we train a state-of-the-art emotion recognition model, and leverage this during training of \model as supervision. 
\model introduces a novel perceptual \textit{emotion consistency loss} that encourages the similarity of emotional content between the input and rendered reconstruction.

While the new emotion consistency loss results in better reconstructed emotions, this alone is insufficient.
Large image datasets used by previous 3D reconstruction methods, while containing a large number of subjects of diverse ethnicities, lack emotional expressivity \cite{Klare2015_IJBA,Cao2018_VGGFace2,Wang_2019_ICCV,Chung2018_VoxCeleb2}. 
Large datasets with facial expressions, valence, and arousal in-the-wild, on the other hand, while rich in emotions, do not provide multiple images per subject in diverse conditions \cite{affectNet, BenitezQuiroz2016_EmotioNet,Aff-Wild2, AFEW-VA, SewaDB, kaisiyuan2020mead, cao14_crema} 
and smaller datasets in controlled settings are not suitable for deep learning \cite{mmi_db, cohn10_ckplus, disfa, disfaplus, belfastemo}.
Multiple  images of the same person, however, are required to train current state-of-the-art 3D face reconstruction methods \cite{Sanyal2019_RingNet,Feng2021_DECA,Deng2019}.
To overcome this, \model builds on top of DECA \cite{Feng2021_DECA}, a publicly available 3D face reconstruction framework that achieves state-of-the-art identity shape reconstruction accuracy \cite{Feng2018evaluation,Sanyal2019_RingNet}. 
Specifically, we augment DECA's architecture with an additional trainable prediction branch for facial expression, while keeping other parts fixed.
This enables us to only train the expression part of \model on emotion-rich image data \cite{affectNet}, which results in improved emotion reconstruction performance, while retaining DECA's identity face shape quality.  

Once trained, \model reconstructs a 3D face from a single image (Fig.~\ref{fig:teaser}), it significantly outperforms previous state-of-the-art methods in terms of the reconstructed expression quality, it preserves the state-of-the-art identity shape reconstruction accuracy, and the reconstructed face can be readily animated. 
Further, the expression parameters regressed by \model convey sufficient information for in-the-wild emotion recognition, with on-par performance with the best image-based methods \cite{toisoul2021estimation}.

In summary, our main contributions are:
1) The first approach to reconstruct an animatable 3D face model from an in-the-wild image, that is capable of recovering facial expressions that convey the correct emotional state. 
2) A novel perceptual emotion-consistency loss that rewards the accuracy of the reconstructed emotion. 
3) The first 3D geometry-based framework for in-the-wild emotion recognition, with comparable performance to current state-of-the-art image-based methods. 
4) The code and model are publicly available for research purposes at \url{https://emoca.is.tue.mpg.de}.

\section{Related work}

\qheading{Monocular face reconstruction:} 
Reconstructing 3D face shape from images has been studied extensively for more than two decades \cite{Egger2020,zollhoefer_survey_2018}.
Model-free approaches 
directly regress 3D meshes \cite{Deng2020_RetinaFace,Dou2017,Feng2018,Guler2017,Jung2021,Ruang2021_SADRNet,Sela2017,Szabo2019,Wei2019,Zeng2019_DF2Net,Wu2020} or voxels \cite{Jackson2017} from an image, or optimize a Signed Distance Function (SDF) \cite{Park2019_DeepSDF} to fit a face image. 
Most of these methods require explicit 3D supervision during training.
While the output is model-free, acquiring the training data typically relies on a 3D face model (3D Morphable Model, or 3DMM).
Thus their ability to reconstruct expressive faces may be limited by the 3DMM-based reconstruction used to generate the paired training data \cite{Deng2020_RetinaFace,Feng2018,Guler2017,Jackson2017,Jung2021,Ruang2021_SADRNet,Wei2019}, the domain gap between 3DMM-based synthetic training data and real images \cite{Dou2017,Sela2017,Zeng2019_DF2Net}, or the regularization towards a fixed 3DMM fitting result \cite{Chatziagapi2021_SIDER}. 
Instead, \model is trained in a self-supervised fashion without any explicit 3D supervision, which enables it to capture less constrained expressions.  
Other self-supervised methods do not leverage face-domain-specific knowledge, which makes them applicable to general objects, but also limits the reconstruction quality \cite{Szabo2019,Wu2020}. 
Unlike \model, none of these model-free methods separate facial identity from facial expression, making them inappropriate for applications like expression re-targeting or animation.

Several works reconstruct the parameters of fixed statistical models like the Basel Face Model (BFM) \cite{bfm09}, FaceWarehouse \cite{Cao2014_FaceWarehouse}, or FLAME \cite{FLAME:SiggraphAsia2017}, or jointly learn a model and reconstruct faces from images \cite{LuanTran2019, Tewari2019_FML, Tewari2018}.
Existing methods can be categorized into optimization-based \cite{AldrianSmith2013,Bas2017fitting,Blanz2002,BlanzVetter1999,Gerig2018,Koizumi2020_UMDFA,Ploumpis2020,RomdhaniVetter2005,Thies2016,VetterBlanz1998} and learning-based.
The latter are trained fully supervised \cite{AnhTran2017,AnhTran2018,Chang2018_ExpNet,Guo2020towards_3DDFA_V2,Kim2018_InverseFaceNet,Richardson2016,Zhu2016_3DDFA} 
or self-supervised with predicted 2D keypoints \cite{Deng2019,Liu2017,Sanyal2019_RingNet,Tewari2017,Tewari2018,Tewari2019_FML,Feng2021_DECA,Shang2020_MGCNET,yang2020facescape}, 
2D face contours \cite{Liu2017}, photometric constraints \cite{Deng2019,Genova2018,Tewari2017,Tewari2018,Tewari2019_FML,Feng2021_DECA,Shang2020_MGCNET,yang2020facescape}, 
face recognition features \cite{Deng2019,Genova2018,Feng2021_DECA,Shang2020_MGCNET}, multi-view constraints \cite{Shang2020_MGCNET}, or multi-image constraints \cite{Deng2019,Genova2018,Sanyal2019_RingNet,Feng2021_DECA,Tewari2019_FML}.
Each supervision signal impacts the reconstructed 3D face in a unique way. 
Explicit 3D mesh or model parameter supervision induces a bias towards the method used to generate the pseudo-ground truth.
Using face recognition features or leveraging multiple images of the same identity during training mainly impacts identity shape and appearance.
Keypoint losses impact the facial geometry and image alignment (global transformation, identity and expression shape parameters), but predicted keypoints are  sparse (commonly 51-68 points), often inaccurate - especially for extreme expressions and head poses - and obtaining the optimal embedding of the corresponding keypoints on the model's surface is challenging.
Photometric losses impact all model parameters (global transformation, identity and expression shape, appearance, and lighting), but,  as with the keypoint losses, are strongly affected by misalignments between the predicted 3D face and the image. 
While using multi-view data during training has the potential to reconstruct more accurate 3D faces, there are no large datasets with a large number of identities and large diversity in expression, ethnicity, age, lighting conditions, etc.
Consequently, while the field of monocular in-the-wild face capture has made tremendous progress, there are still limitations, particularly in the accuracy of the reconstructed expressions, which limit the emotions that can be perceived from the reconstructed 3D shapes. 
\model instead learns to reconstruct expressive faces by combining emotion features that mainly propagate to the reconstructed expression, with a unique self-supervised framework that enables us to leverage a large dataset of diverse expressions.

\qheading{Emotion analysis from images:}
Emotion analysis is a long-standing problem in computer vision and related fields (see \cite{oxfordHandbook, ALAMEDAPINEDA20191} for comprehensive surveys). 
Emotional states are commonly represented as discrete basic \cite{Ekman1971constants,Ekman1992argument} (e.g., Happiness, Surprise, ...) or compound expression categories \cite{Du2014compound} (e.g., happily surprised), continuous valence (positive-negative) and arousal (relaxed-intensive) values \cite{russell1980circumplex}, or Facial Action Units (FACS) activations \cite{EkmanFriesen1978_FACS}, where each action unit (AU) corresponds to a particular emotion-related facial muscle movement. 

Early work on expression recognition extracts geometric features defining shape and location of face components \cite{Tian2001recognizing,PanticRothkrantz}, appearance features \cite{Feng2005_LBP,Shan2009_LBP}, or combinations of these \cite[Chapter 19]{Jain2011handbook}. 
Over the last decade, the availability of large datasets for 
single-image expression analysis \cite{affectNet,BenitezQuiroz2016_EmotioNet} and audio-visual videos \cite{Aff-Wild2, AFEW-VA, SewaDB} shifted the focus from manually designed features to end-to-end trained models \cite{Li2020}. 
While early work like Wen and Huang~\cite{WenHuang2003} uses 3D non-rigid surface tracking to extract features for expression reconstruction, the majority of 3D-based methods focus on recognizing expressions from 3D scans \cite{Nonis2019_survey,Sandbach2012_survey}.
Among these, the most relevant to \model, is
\cite{Ramanathan2006} 
as they use 3DMM features to classify three expressions (obtained by fitting the 3DMM to the scans); most other methods use diverse 2D and 3D features extracted from the textured 3D scans. 

Few 3DMM-based methods exist to recognize expressions from images. 
Bejaoui et al.~\cite{Bejaoui2017} fit a 3DMM to images, while Chang et al.~\cite{Chang2018_ExpNet} and Koujan et al.~\cite{Koujan2020} train a 3DMM parameter regressor, fully-supervised by parameters obtained by fitting a 3DMM to images and videos. 
From the 3DMM expression parameters, they then learn to classify different expressions. 
Most related to \model, Shi et al.~\cite{Shi2020_3DMM_ExpRecon} use an expression recognition loss during training, but with the goal of obtaining a more discriminative latent representation. 
These methods focus on recognizing expressions, not improving 3D reconstruction.
In contrast, \model leverages recent advances in emotion recognition to reconstruct more expressive 3D faces.

\section{Preliminaries}
\label{sec:preliminaries}

\qheading{Face model:}
FLAME \cite{FLAME:SiggraphAsia2017} is a statistical 3D head model with parameters for identity shape $\shapecoeff \in \mathbb{R}^\shapedim$, facial expression $\expcoeff \in \mathbb{R}^\expdim$, and pose parameters $\posecoeff \in \mathbb{R}^{3\numjoints+3}$ for rotations around $\numjoints = 4$ joints (neck, jaw, and eyeballs) and the global rotation.
Given all parameters, FLAME outputs a mesh with $\numverts=5023$ vertices. 
Formally, FLAME is:
\begin{equation}
 \flamev(\shapecoeff, \posecoeff, \expcoeff)\rightarrow (\mathbf{V}, \mathbf{F}),
\end{equation}
with vertices $\mathbf{V} \in \mathbb{R}^{\numverts \times 3}$ and $\numfaces = 9976$ faces $\mathbf{F} \in \mathbb{R}^{\numfaces \times 3}$. 
FLAME comes with an appearance model, converted from Basel Face Model's 
albedo space \cite{bfm09} to FLAME's UV layout \cite{BFM_to_FLAME}. 
Given parameters $\albedocoeffs \in \albedospaceexpl$, this model outputs a FLAME texture map $\albedo(\albedocoeffs) \in \mathbb{R}^{d \times d \times 3}$.

\qheading{Face reconstruction:}
DECA \cite{Feng2021_DECA} is a publicly available framework to reconstruct a detailed, animatable 3D face model from a single image. 
We follow DECA's notation for simplicity. %
Given an image $\image$, the coarse encoder 
\begin{equation}
    \coarseencoder(\image) \rightarrow (\shapecoeff, \posecoeff, \expcoeff, \albedocoeffs, \lighting, \cam)
    \label{eq:coarseencoder}
\end{equation}
outputs FLAME geometry parameters $\shapecoeff, \posecoeff, \expcoeff$, albedo $\albedocoeffs$, Spherical Harmonics (SH) \cite{Ramamoorthi2001_SH} lighting $\lighting \in \mathbb{R}^{27}$, and camera $\cam \in \mathbb{R}^3$, which is the concatenation of isotropic scale $s \in \mathbb{R}$ and translation $\textbf{t} \in \mathbb{R}^2$.
The detail encoder 
\begin{equation}
    \detailencoder(\image) \rightarrow \zcode
\end{equation}
encodes $\image$ to a subject-specific detail vector $\zcode \in \mathbb{R}^{128}$.
To reconstruct dynamic expression wrinkles, the detail decoder
\begin{equation}
    \edm(\zcode, \expcoeff, \posecoeff_{jaw}) \rightarrow  \displacements
\end{equation}
uses $\zcode$ to parametrize static person-specific details, and FLAME's expression $\expcoeff$ and jaw-pose parameters $\posecoeff_{jaw}$ to generate an expression-dependent detail UV displacement map $\displacements \in \mathbb{R}^{d \times d \times 3}$. 

Denoting the rendering function with $\renderer$ \cite{Ravi2020_PyTorch3D}, the coarse shape can be rendered to a 2D image as $\renderer( \flamev(\shapecoeff, \posecoeff, \expcoeff), \albedocoeffs, \lighting, \cam )\rightarrow \image_{Rc}$.
To render the FLAME mesh, with expression-dependent details, to an image $\image_{Rd}$, the $\displacements$ are converted to a detailed normal map $ \normals_d$, and provided as additional parameters to $\renderer$; formally $\renderer( \flamev(\shapecoeff, \posecoeff, \expcoeff), \albedocoeffs, \lighting, \cam, \normals_d  )\rightarrow \image_{Rd}$. 

\qheading{Relative keypoint loss:}
Given 2D face keypoints $\landmark_i \in \mathbb{R}^2$ and the corresponding keypoints on FLAME's mesh surface $\flamev_i \in \mathbb{R}^3$, the relative keypoint loss \cite{Feng2021_DECA} computes offset vectors between pairs of 2D keypoints and between the corresponding pairs of projected model keypoints, and penalizes the difference. 
Formally, the loss computes as 
\begin{equation}
    L_{\mathit{rk}}^{E} = \sum \limits_{(i,j) \in E} \norm{\landmark_i - \landmark_j - s\Pi(\flamev_i - \flamev_j)}_1,
\end{equation}
where $E$ is a set of landmark index pairs, and $\Pi \in \mathbb{R}^{2 \times 3}$ is the orthographic 3D-2D projection matrix. 

\qheading{Emotion recognition:}
For the emotion network, we use ResNet-50 as the backbone,  with a fully connected prediction head that outputs expression classification, valence, and arousal. See Appendix for experiments with other backbones.
The network is trained on AffectNet \cite{affectNet}, a large-scale annotated emotion dataset. 
We adapt the training setting from Toisoul et al.~\cite{toisoul2021estimation} with minor modifications as described in the Appendix.
The loss function consists of several terms such as categorical cross entropy for expression classification, mean squared error and correlation coefficient losses for valence and arousal; see Appendix for details of the losses.
After the network is trained, prediction heads are discarded, and the features of the final layer of the backbone network serve as our emotion feature $\emovec \in \mathbb{R}^{\left| \emovec \right|}$.
We denote the emotion network as $\emonet(\image) \rightarrow \emovec$.

\section{Method: EMOCA}
\label{sec:method}

The main goal of \model is to address a significant limitation of the prior art - to recover 3D face shapes from single images that convey the full spectrum of emotion.
Our technical contribution is twofold, first, we introduce a novel \textit{emotion consistency loss} that is designed to encourage \textit{emotion similarity} between the input image and the output rendering as training supervision.
Second, we leverage parts of DECA's \cite{Feng2021_DECA} trained model in order to only train the expression part of \model on emotion-rich image data, while preserving DECA's identity shape reconstruction performance. 

\qheading{Architecture:}
\model's architecture is based on DECA \cite{Feng2021_DECA}.
As with many state-of-the-art methods, DECA takes an input image and uses several neural networks to factor it into shape, albedo, lighting, etc.
Given these factors, one can differentiably render an output image that should look like the input.
Here we exploit this output image in a novel way by encouraging it to have the same {\em expression} as the input image.

Training models like DECA \cite{Feng2021_DECA} on emotion-rich image data \cite{affectNet} is infeasible, due to DECA's requirement of multiple training images per subject to regularize the training of the identity shape reconstruction of $\coarseencoder$ (Eq.~\ref{eq:coarseencoder}). 
Instead, we augment DECA's architecture with an additional expression encoder
\begin{equation}
    \coarseencoderexp(\image) \rightarrow \expcoeff_e, 
\end{equation}
and keep the weights of $\coarseencoder$ fixed during training,  thereby retaining the predictions of $\shapecoeff, \posecoeff, \albedocoeffs, \lighting $ and $ \cam$ from DECA, but discarding DECA's $\expcoeff$. 
Further, let $\renderer( \flamev(\shapecoeff, \posecoeff, \expcoeff_e), \albedocoeffs, \lighting, \cam )\rightarrow \image_{Re}$ denote the rendering of the output of $\coarseencoder$ with the expression of the input image, $\coarseencoderexp(\image)$.

\begin{figure*}[t]
    \offinterlineskip
    \centering
    \includegraphics[width=2.0\columnwidth]{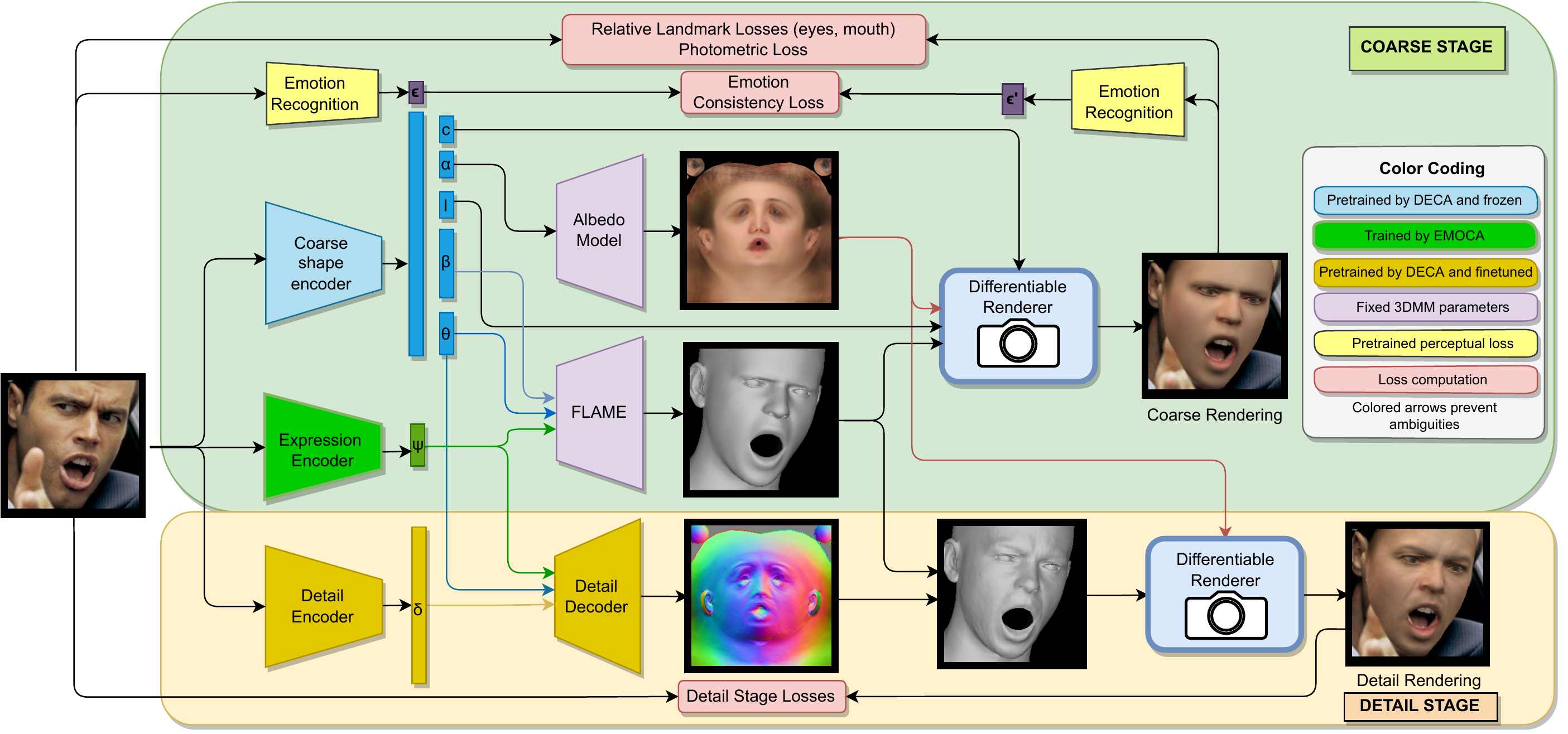}
    \caption{
        \model overview.
        For the coarse training stage (green box), the input image is fed to the coarse shape encoder (initialized from DECA \cite{Feng2021_DECA} and fixed) and \model's trainable expression shape encoder. 
        A textured 3D mesh is then reconstructed from the regressed identity shape, expression shape, pose, and albedo parameters with FLAME's geometry and albedo models as fixed decoders. 
        This textured mesh is rendered by a differentiable renderer with the regressed camera and spherical harmonics lighting.
        Our novel emotion consistency loss (Eq.~\ref{eq:emo}) penalizes the difference between the emotion features of the input image and those of the rendered coarse shape, after passing both images through a fixed emotion recognition network. 
        For the detail training stage (yellow box), \model's expression encoder is fixed, and the regressed expression (and jaw-pose) parameters are used to condition the detail decoder. 
    }
    \label{fig:overview}
\end{figure*}

For an overview of the model architecture, see Figure~\ref{fig:overview}.
Training $\coarseencoderexp$ only has several advantages:
1) Training datasets are not required to contain multiple images per subject.
2) Not training the identity prediction enables us to remove the face recognition loss.
3) Having fixed pose, shape, and camera parameters allows us to remove the landmark reprojection loss. 
4) This results in reduced training resources, faster training time, and reduced memory consumption due to the lower number of training parameters. 

\qheading{Loss function:}
In total, we optimize: 
\begin{eqnarray}
    L & = &  \lambda_{emo} L_{\mathit{emo}} + \lambda_{pho} L_{\mathit{pho}}  
    + \lambda_{eye} L_{\mathit{eye}} \nonumber\\
    & &  + \ \lambda_{mc} L_{\mathit{mc}} + \lambda_{lc} L_{\mathit{lc}} + \lambda_{\expcoeff} L_{\expcoeff}
\end{eqnarray}
with emotion consistency loss $L_{\mathit{emo}}$, photometric loss $L_{\mathit{pho}}$, eye closure loss $L_{\mathit{eye}}$, mouth closure loss $L_{\mathit{mc}}$, lip corner loss $L_{\mathit{lc}}$, and expression regularizer $L_{\expcoeff}$, each weighted by a factor $\lambda_{x}$.

\qheading{Emotion consistency loss:} 
The emotion consistency loss computes the difference between the emotion features of the input image $\emovec_{\image} = \emonet(\image)$ and those of the rendered image, $\emovec_{Re} = \emonet(\image_{Re})$ as:
\begin{equation}
    L_{\mathit{emo}} = \emometric(\emovec_{\image}, \emovec_{Re}),
    \label{eq:emo}
\end{equation}
with $\emometric(\emovec_1, \emovec_2) = \norm{ \emovec_1  - \emovec_2 }_2$.
Instead of measuring a geometric error, $L_{\mathit{emo}}$ computes a perceptual difference between the input image and the rendered image. 
Optimizing this loss during training ensures that the reconstructed 3D face conveys the emotional content of the input image. 

\qheading{Photometric loss:}
The photometric loss computes the pixel error between the input image $\image$ and the output rendering $\image_{Re}$.
$L_{\mathit{pho}} = \norm{V_{\image} \odot (\image - \image_{Re})}_{1,1}.$
$V_{\image}$ denotes a rendered mask of the output face shape, with each pixel located in the face skin region is equal to $1$, and $0$ elsewhere. 
The operator $\odot$ denotes the Hadamard product.

\qheading{Eye closure loss:}
The eye closure loss computes as $L_{\mathit{eye}} = L_{\mathit{rk}}^{E_{\mathit{eye}}}$, where $E_{\mathit{eye}}$ is a set of upper/lower eyelid keypoint pairs. 
Due to slight misalignment between image landmarks and projected 3D landmarks, enforcing standard landmark reprojection losses produces incorrect predictions. 
Instead, using (translation-invariant) relative keypoint losses (for eye closure, mouth closure, and mouth width) is less susceptible to misalignments. 

\qheading{Mouth closure loss:}
The loss computes as $L_{\mathit{mc}} = L_{\mathit{rk}}^{E_{\mathit{mc}}}$, where $E_{\mathit{mc}}$ is a set of upper/lower lip keypoint pairs. 

\qheading{Lip corner loss:}
The lip corner loss computes as $L_{\mathit{lc}} = L_{\mathit{rk}}^{E_{\mathit{lc}}}$, where $E_{\mathit{lc}}$ is the pair of left and right lip corners. 

\qheading{Expression regularization:}
The expression is regularized as $L_{\expcoeff} = \norm{\expcoeff}_2^2$.

\qheading{Detailed stage:}
The detail training stage adds wrinkle details that are animatable.
Here we follow DECA's design, and use the same architecture and losses. 

\section{Experiments}
\label{sec:experiments}

\begin{table*}[t]
\centering
\resizebox{0.9\textwidth}{!}{

\begin{tabular}{l|cccc|cccc|c}
\toprule
                   Model &  V-PCC $\uparrow$ &  V-CCC $\uparrow$ &  V-RMSE $\downarrow$ &  V-SAGR $\uparrow$ &  A-PCC $\uparrow$ &  A-CCC $\uparrow$ &  A-RMSE $\downarrow$ &  A-SAGR $\uparrow$ &  E-ACC $\uparrow$ \\
\midrule
 EmoNet \cite{toisoul2021estimation} &   0.75 &   0.73 &    \emph{0.32} &    \emph{0.80} &   0.68 &   0.65 &    \textbf{0.29} &    \emph{0.78} &      \emph{0.68} \\ \hline
          Deep3DFace \cite{Deng2019} &   0.75 &   0.73 &    0.33 & \emph{0.80} &   0.66 &   0.65 &    0.31 &    0.78 &      0.65 \\ 
        ExpNet \cite{Chang2018_ExpNet} &   0.45 &   0.42 &    0.43 &    0.73 &   0.39 &   0.36 &    0.38 &    0.64 &      0.46 \\
            MGCNet \cite{Shang2020_MGCNET}  &   0.71 &   0.69 &    0.35 &    \emph{0.80} &   0.59 &   0.58 &    0.34 &    0.77 &      0.60 \\
           3DDFA\_V2 \cite{Guo2020towards_3DDFA_V2}  &   0.63 &   0.62 &    0.39 &    0.75 &   0.53 &   0.50 &    0.34 &    0.73 &      0.52 \\
            DECA \cite{Feng2021_DECA} &   0.70 &   0.69 &    0.36 &    0.76 &   0.59 &   0.58 &    0.33 &    0.74 &      0.59 \\
               DECA w/ details \cite{Feng2021_DECA} &   0.70 &   0.69 &    0.37 &    0.77 &   0.59 &   0.57 &    0.33 &    0.77 &      0.58 \\ \hline
             EMOCA (Ours) &   \textbf{0.78} &   \textbf{0.77} &    \textbf{0.31} &    \textbf{0.81} &   \emph{0.69} &   \emph{0.68} &    \emph{0.30} &    \emph{0.81} &      \emph{0.68} \\
  EMOCA w/ details (Ours) &   \emph{0.77} &   \emph{0.76} &    \textbf{0.31} &    \textbf{0.81} &   \textbf{0.70} &   \textbf{0.69} &    \textbf{0.29} &    \textbf{0.83} &      \textbf{0.69} \\

\bottomrule
\end{tabular}}
\caption{
{\bf Emotion recognition performance on the AffectNet test set \cite{affectNet}.}
The EmoNet performance is measured using the model that is publicly released by the authors. 
For \model and the other 3D baselines, we train the recognition module as described in \secref{para:er}. DECA w/ detail means that DECA's detail code prediction was included in the input to the regressor, along with the 3DMM parameters.
Please note that \model's performance is on par with EmoNet and it outperforms all other 3D reconstruction-based methods. 
}
\label{table:emorec}
\end{table*}

\begin{figure*}[t]
    \offinterlineskip
    \centering
    \includegraphics[width=1.025\columnwidth]{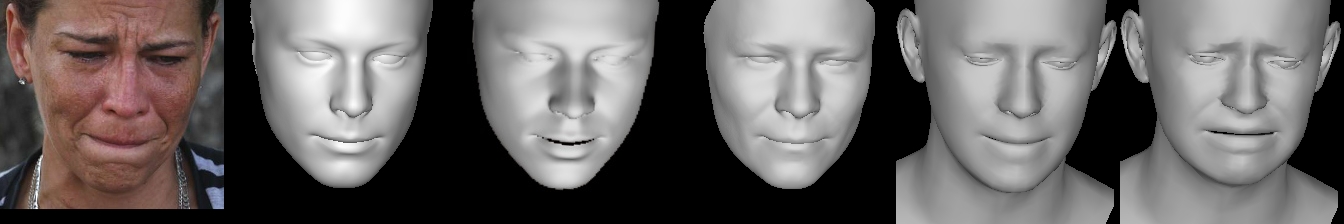} 
    \includegraphics[width=1.025\columnwidth]{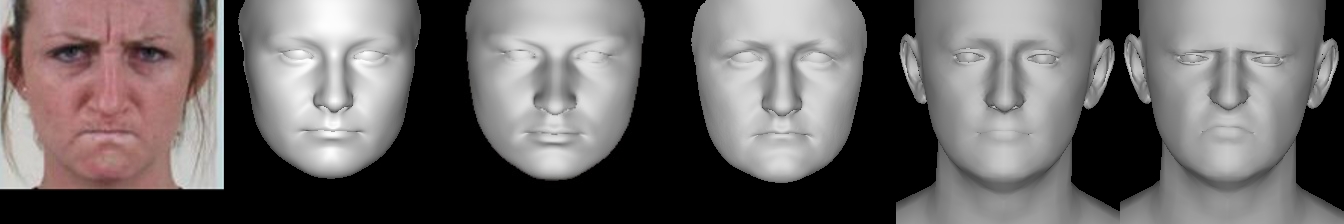} \\
    \includegraphics[width=1.025\columnwidth]{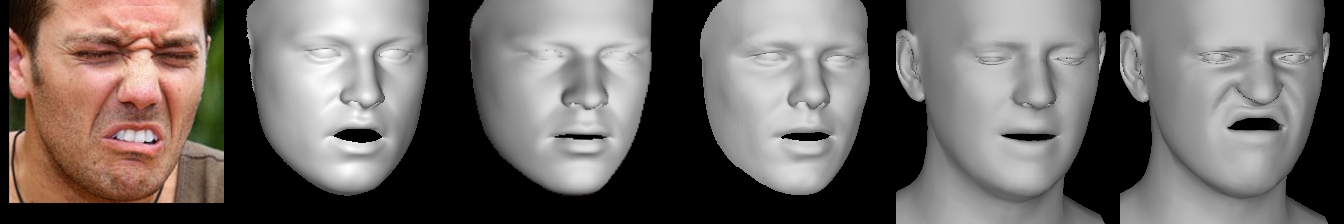} 
    \includegraphics[width=1.025\columnwidth]{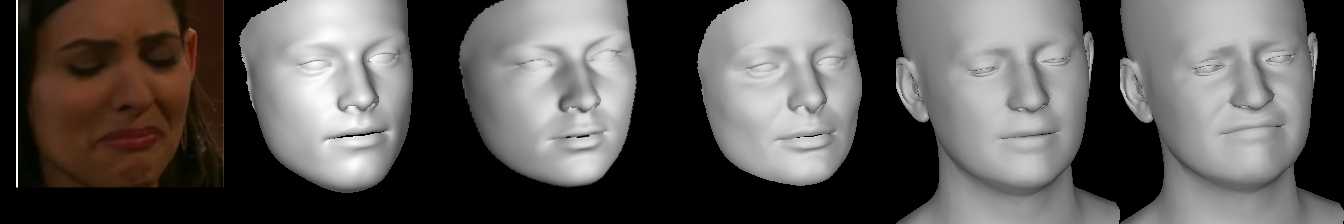} \\
    \includegraphics[width=1.025\columnwidth]{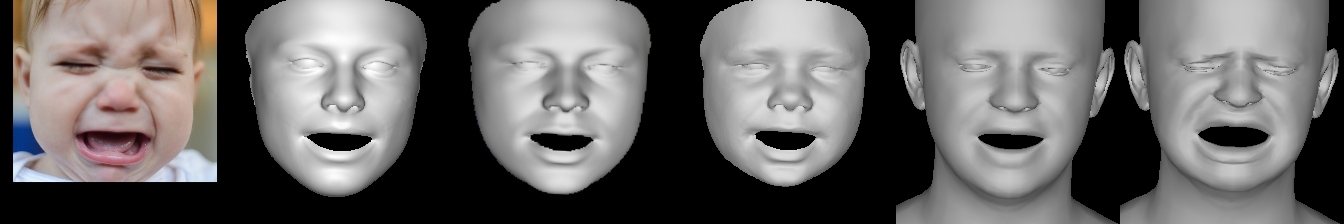} 
    \includegraphics[width=1.025\columnwidth]{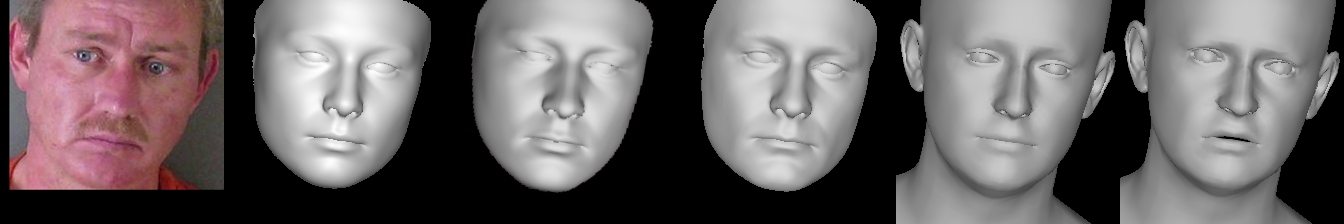} \\
    \includegraphics[width=1.025\columnwidth]{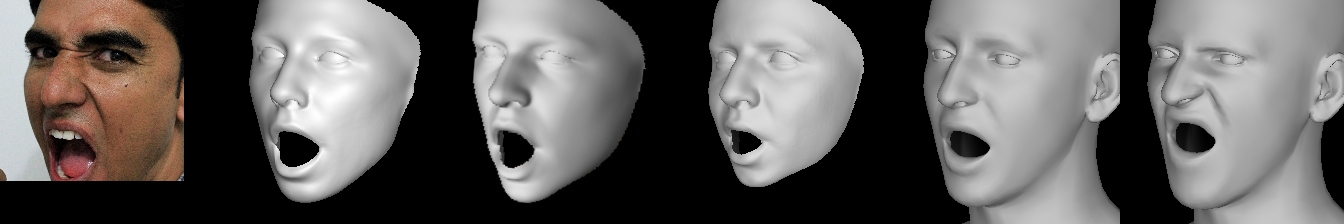} 
    \includegraphics[width=1.025\columnwidth]{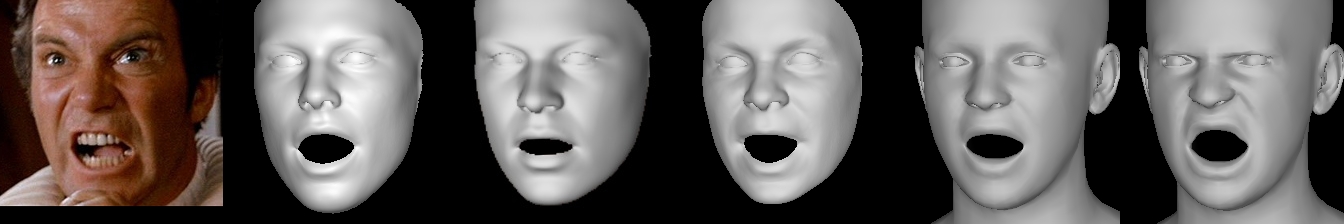} \\
    \includegraphics[width=1.025\columnwidth]{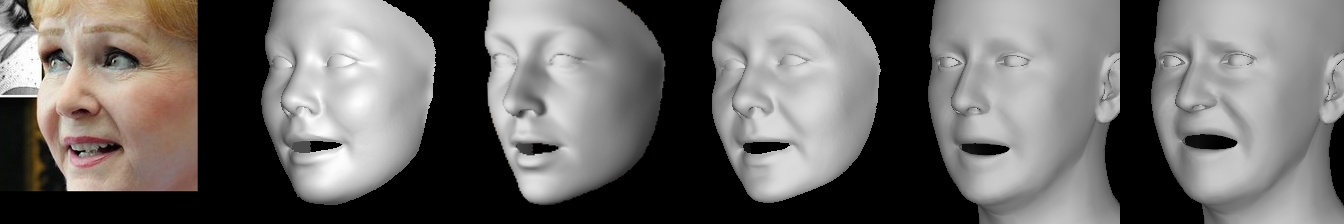} 
    \includegraphics[width=1.025\columnwidth]{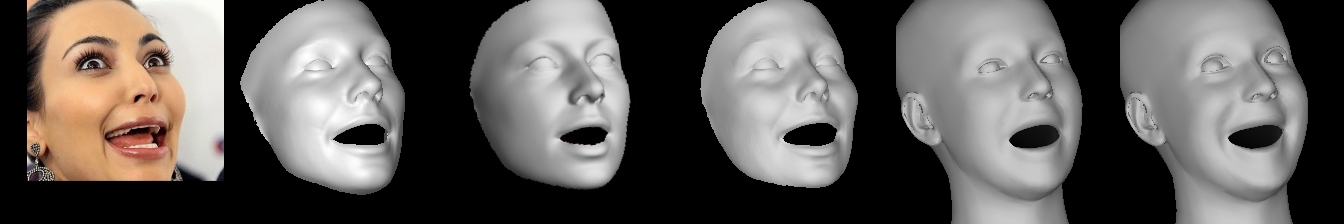} \\
    \includegraphics[width=1.025\columnwidth]{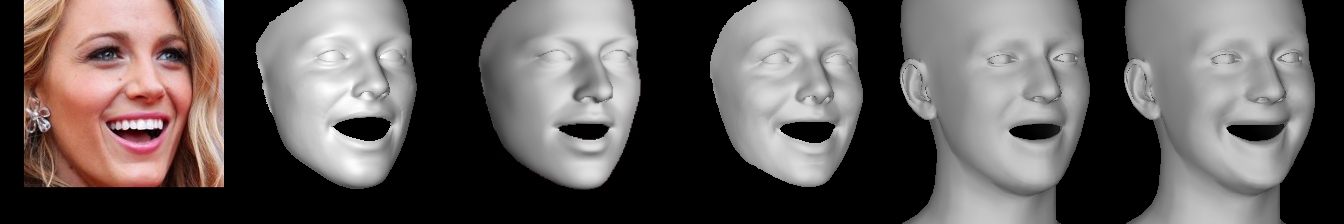} 
    \includegraphics[width=1.025\columnwidth]{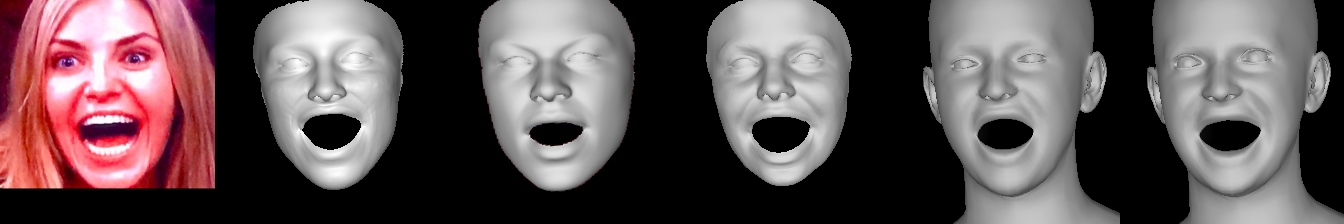} 
    \caption{Comparison of {\bf coarse reconstruction} methods, from left to right: Input, 3DDFA\_V2 \cite{Guo2020towards_3DDFA_V2}, MGCNet \cite{Shang2020_MGCNET},  Deng et al.~\cite{Deng2019}, DECA \cite{Feng2021_DECA} (coarse), and \model (coarse). 
    \model conveys the emotions of the input images better than other methods.
    }
    \label{fig:qualitative1}
\end{figure*}

\subsection{Training setting}
The first stage (coarse part) of \model is trained with AffectNet \cite{affectNet} for a maximum of 20 epochs, with early stopping, using the Adam optimizer \cite{Kingma2015} and a learning rate of $5e-5$.
We use the same training/validation/testing split as proposed by \cite{toisoul2021estimation}. We set $\lambda_{emo}=1$, $\lambda_{pho}=2$, $\lambda_{eye} = \lambda_{lc} = \lambda_{mc} =0.5$ and $\lambda_{\expcoeff}=1e-4$. 
\model's second stage (detail part) training is comparable to DECA's second stage training. 
We use the same training data \cite{Chung2018_VoxCeleb2, Cao2018_VGGFace2} and train with the same settings. 
Please refer to the Appendix for more training details.

\subsection{Quantitative evaluation}
While, for the task of 3D face reconstruction, standard benchmarks exist to quantitatively evaluate the identity face shape \cite{Feng2018evaluation,Sanyal2019_RingNet}, no such benchmark exists to assess the accuracy of the reconstructed expression. 
Unlike the identity shape benchmarks, quantitatively measuring the difference between a reconstructed 3D facial expression and a ground truth scan is less meaningful.
The errors would be dominated by errors of the reconstructed identity face shape, and a low geometric error would not necessarily correspond to a small difference in human perception of the emotion.  
Instead, we evaluate \model 1) qualitatively, 2) quantitatively for the task of in-the-wild emotion recognition, and 3) perceptually in an Amazon Mechanical Turk (AMT) study. 

\qheading{Emotion recognition:}
\label{para:er}
Our goal is to quantify how much of the input emotion is conveyed in the reconstructed 3D face. 
For this, we apply 3D face reconstruction methods to in-the-wild emotional face images and evaluate emotion recognition accuracy based on the 3D reconstruction.
Here we focus on methods that reconstruct a parametric model of the face, i.e.~a 3DMM.
To that end, for each 3D face reconstruction method, we train a 4-layer MLP with Batch Normalization \cite{IoffeS15} and LeakyReLUs to regress valence and arousal levels, and classify expression labels directly from the predicted 3DMM parameters. 
The training details are described in the Appendix.

We evaluate emotion recognition on the AffectNet test set \cite{affectNet} and the AFEW-VA test set \cite{AFEW-VA}.
For each method, we report Concordance correlation coefficients (CCC $\uparrow$), Pearson correlation coefficients (PCC $\uparrow$), root mean squared error (RMSE $\downarrow$), and sign agreement (SAGR $\uparrow$) for valence (V) and arousal (A) regression and accuracy for expression (E) classification on the test set defined by \cite{toisoul2021estimation}.
\model outperforms all 3D face reconstruction methods, and is on par with the image-based state-of-the-art \cite{toisoul2021estimation}.  
For details, see \tabref{table:emorec} for results on the AffectNet dataset and \tabref{table:affewvatest} of the Appendix for the AFEW-VA dataset.

Note that \model performs on par with EmoNet \cite{toisoul2021estimation}, which is a recent method for estimating emotion from images.
This confirms that the emotional content is present in our 3D reconstruction and that 3D shape is sufficient to understand emotion.
This has implications for future research on emotion recognition.

\qheading{Perceptual study:}
The 3D geometry reconstructed from an image must convey the emotion of the input image. 
Directly comparing rendered geometry with an image is difficult due to the domain gap. 
Instead, we perform a perceptual study using AMT to assess the perceived expression of emotion from rendered 3D reconstructions. 
Specifically, given an image, we ask participants to categorize the perceived expression of emotion into one of the 7 basic emotions (Anger, Disgust, Fear, Happiness, Sadness, Surprise, and Contempt) or as a neutral expression (no emotion).
A single evaluation task contains 75 images in random order; 35 real images, the 35 corresponding rendered reconstructions (from one method), and 5 qualification samples.
The 5 qualification samples are duplicates sampled from the 35 real images, and they are chosen to be of easily recognizable emotion.
Each task is performed by 10 participants. 
Participants that either misclassify the emotion, or inconsistently label the duplicated images, for at least 2 (of the 5) qualification samples are discarded from further analysis to filter out inattentive and/or uncooperative participants.
For each method, we measure the classification consistency between each participant's labels for the rendered images and their labels for the corresponding real images. 
If the rendered 3D meshes contain the emotional content of the images, then the scores given to both should be consistent.

We select 35 images with balanced emotional content (i.e., 5 images per basic emotion) from the AffectNet test set \cite{affectNet}.
For each image, we reconstruct 3D faces using \model, DECA \cite{Feng2021_DECA}, Deep3DFace \cite{Deng2019}, MGCNet \cite{Shang2020_MGCNET} and 3DDFA\_V2 \cite{Guo2020towards_3DDFA_V2}.
The classification consistency averaged across participants for each method are: \model (coarse) \textbf{$0.68$}, \model (detail) \emph{$0.65$}, Deep3DFace $0.37$, DECA (coarse) $0.33$, DECA (detail) $0.31$, MGCNet $0.32$, 3DDFA\_V2 $0.31$.
In summary, \model preserves the emotional content of images better than the other methods. 
Note that, perhaps surprisingly, there is little difference between the scores for the coarse meshes of \model/DECA and the detailed ones. 
Despite more wrinkle detail, our perceptual experiments suggest that the detailed meshes do not convey more emotional content.
One possible explanation is that, in addition to adding valid wrinkle details,  the detail generator sometimes adds artifacts in the lip region (e.g.~\figref{fig:teaser}, col.~1 \& 3), and hallucinates details in the forehead (e.g. \figref{fig:teaser}, col.~8).
These could negatively impact participants' perception. 
For the full confusion tables, see the Appendix.

\qheading{Emotion recognition vs.~perceptual study:}
There is a considerable discrepancy between the results of the automatic emotion recognition results and the perceptual study results, in particular for Deep3DFace \cite{Deng2019}.
Deep3DFace performs much better on the emotion recognition task (slightly below SOTA), than on the perceptual study. 
Unlike EMOCA, it is not capable of producing highly emotional reconstructions (see \figref{fig:qualitative1}). 
We hypothesize that the automatic predictors are capable of detecting more subtle cues than humans. 
We investigate this by measuring the agreement (i.e., percentage of matching predictions) between the method's classifier (from the reconstructed face parameters) and the participant's annotation of the \textit{input images} from the perceptual study. 
The results are: EMOCA $62\%$ and Deep3DFace $62\%$. 
This indicates that the predicted parameters for both methods contain a similar amount of information about the emotions compared to the annotations of the input images.
However, the agreement between the method's classifier and the participant's annotation of the \textit{rendered reconstructions} is for EMOCA $48\%$, and for Deep3DFace $26\%$.
In other words, EMOCA is signficantly more in agreement with human perception.

\subsection{Qualitative evaluation}
\begin{figure}[t]
    \offinterlineskip
    \centering
    \includegraphics[width=0.155\columnwidth]{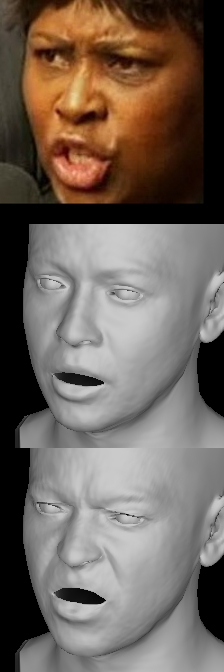} 
    \includegraphics[width=0.155\columnwidth]{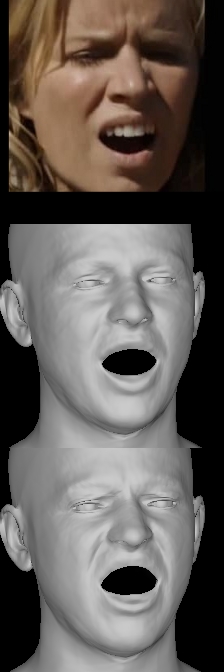} 
    \includegraphics[width=0.155\columnwidth]{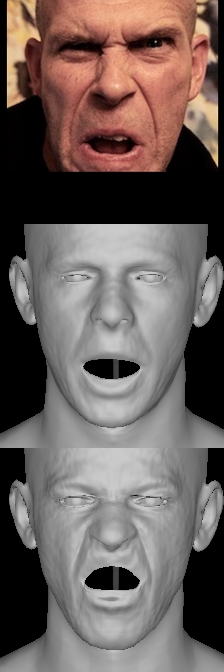} 
    \includegraphics[width=0.155\columnwidth]{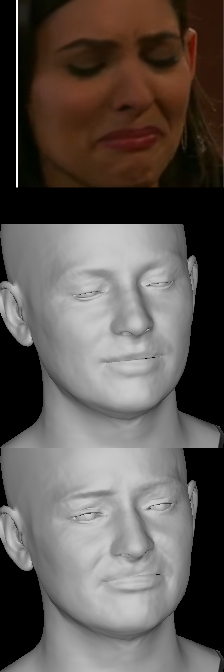} 
    \includegraphics[width=0.155\columnwidth]{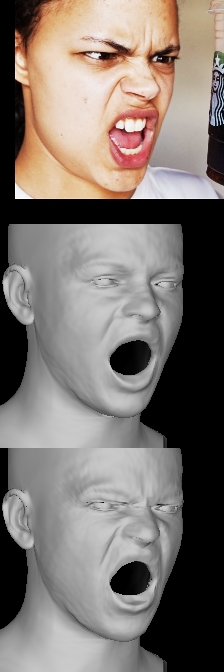} 
    \includegraphics[width=0.155\columnwidth]{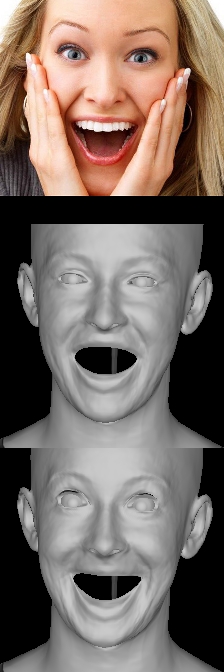} 
    \caption{Comparison of 3D reconstructions with {\bf detail displacements}. Top: Input, Middle: DECA \cite{Feng2021_DECA}, Bottom: \model.
    \model results contain more expression-dependent details that better convey the emotion of the input images than DECA.
    }
    \label{fig:qualitative2}
\end{figure}

We provide a visual comparison of the coarse shape reconstruction methods in Fig.~\ref{fig:qualitative1}. 
Observe that \model outperforms all the previous methods in terms of capturing the emotional content of the original image in the reconstructed expression. 
In \figref{fig:qualitative2} we compare our detail reconstructions to DECA's detail reconstruction. 
Compared to DECA, our detailed displacement better captures the fine details of highly emotional input images.

\subsection{Ablation experiment}

\begin{table*}[t]
\centering
\resizebox{0.9\textwidth}{!}{

\begin{tabular}{l|cccc||cccc|c}
\toprule
                   Model &  V-PCC $\uparrow$ &  V-CCC $\uparrow$ &  V-RMSE $\downarrow$ &  V-SAGR $\uparrow$ &  A-PCC $\uparrow$ &  A-CCC $\uparrow$ &  A-RMSE $\downarrow$ &  A-SAGR $\uparrow$ &  E-ACC $\uparrow$ \\
\midrule
                DECA \cite{Feng2021_DECA} &   0.70 &   0.69 &    0.36 &    0.76 &   0.59 &   0.58 &    0.33 &    0.74 &      0.59 \\
                EMOCA DS w/o Emo &   0.70 &   0.69 &    0.37 &    0.78 &   0.61 &   0.58 &    0.32 &    \emph{0.79} &      0.60 \\
                EMOCA w/o Emo &   0.68 &   0.66 &    0.36 &    0.74 &   0.59 &   0.58 &    0.32 &    0.77 &      0.59 \\
                EMOCA DS &   \emph{0.77} &   \emph{0.76} &    \textbf{0.31} &    \textbf{0.82} &   \textbf{0.69} &   \emph{0.67} &    \textbf{0.29} &    \emph{0.79} &      \textbf{0.68} \\
                EMOCA &   \textbf{0.78} &   \textbf{0.77} &    \textbf{0.31} &    \emph{0.81} &   \textbf{0.69} &   \textbf{0.68} &    \emph{0.30} &    \textbf{0.81} &      \textbf{0.68} \\ 
\bottomrule
\end{tabular}
}
\caption{{\bf Ablation experiment.}
Effect of ablating the data and the emotion consistency loss on EMOCA evaluated on the emotion recognition task.
From top to bottom, we see the performance of DECA, EMOCA trained on the DECA dataset w/o emotion loss, \model w/o emotion loss, \model trained on the DECA dataset, and \model. 
We refer to DECA's training data as DECA dataset (DS), which is a combination of 
VGGFace2 \cite{Cao2018_VGGFace2}, 
and VoxCeleb2 \cite{Chung2018_VoxCeleb2}.
The key finding is that novel emotion consistency loss is critical for the performance of this task, as with emotion loss, \model's performance improves.
Finetuning on AffectNet, which has a much richer variety of facial expressions, only marginally increases in performance over training on DECA's original training data (DS). 
}
\label{table:ablation}
\end{table*}

Table \ref{table:ablation} shows the effect of ablating the training data and the emotion consistency loss. 
The table summarizes the effect of \model trained w/ and w/o the emotion consistency loss, and using the DECA data only \cite{Chung2018_VoxCeleb2, Cao2018_VGGFace2} instead of the AffectNet training data \cite{affectNet}. 

\section{Discussion and limitations}

\qheading{Baseline:}
\model builds on top of DECA due to its state-of-the-art identity shape reconstruction performance. %
We found in our experiments that the recently released Deep3DFaceRecon \cite{Deep3DFaceRecon_PyTorch} gives better 3D face reconstructions than reported in the paper \cite{Deng2019}, and in some cases, it outperforms DECA in terms of the reconstructed expression. 
Combining our emotion consistency loss with the Deep3DFaceRecon framework to further improve their reconstructed expressions is worth further investigation. 

\qheading{Image alignment:}
DECA sometimes predicts 3D faces that are slightly misaligned with the input images. 
\model inherits this limitation due to the fixed coarse shape encoder. 
Further, while \model reconstructs more expressive faces that better convey the emotion of the input image, expressions are also sometimes misaligned.
Mitigating these artifacts, and better balancing the trade-off between geometric alignment and emotion similarity, requires further work. 

\qheading{Emotion embedding analysis:}
We assume that the emotion embedding extracted by the emotion recognition network has desirable properties to guide the optimization of FLAME's expression parameters. 
We found that the emotion recognition loss is more difficult to optimize and it requires more careful weighting of the loss compared to the identity recognition losses used in previous work \cite{Deng2019,Feng2021_DECA,Genova2018}. 
Directly using the pre-trained EmoNet \cite{toisoul2021estimation} for instance did not provide sufficient supervision. 
However, our work is the first to demonstrate how to use emotion recognition features to guide the task of 3D geometry reconstruction. 
In addition using our emotion consistency loss to train \model, we have experimented with the applicability of emotion features for the tasks of emotion retrieval and emotion retargetting via FLAME expression parameter optimization (see Appendix).

\qheading{Emotion network architecture:}
Using a pre-trained state-of-the-art emotion recognition network \cite{toisoul2021estimation} does not provide satisfactory supervision during optimization or training. 
Instead, it produces strong artifacts in the reconstructed geometry. 
To overcome this, we investigate different ResNet \cite{He2016_ResNet} and Swin Transformer \cite{Liu2021_SwinTransformer} based emotion network architectures, and show the effect of different networks in the Appendix.
Based on this analysis, we use a ResNet-50  backbone for our emotion network.

\qheading{Jaw rotations:}
While FLAME's jaw rotation parameters $\posecoeff_{jaw}$ contribute to facial expressions, we found the optimization of $\posecoeff_{jaw}$
to be unstable while training \model.
We hypothesize, that this is due to the lack of a good prior for the jaw rotation. 
However, using different simplified priors for the jaw pose like a simple L2 regularizer did not give satisfactory results. 
We offer a more detailed discussion in the Appendix. 
Investigating the effect of more advanced data-driven jaw priors when optimizing the emotion loss is subject to future work. 

\qheading{Implementation details:}
For details on all hyper parameters and discussion on design choices see the Appendix.

\section{Conclusions}
We have presented \model, a method that takes a single in-the-wild image and reconstructs a 3D face with sufficient facial expression detail to convey the emotional state of the input image. 
\model is trained in a self-supervised fashion from a large dataset of emotion-rich images. 
A novel \textit{emotion similarity} loss provides supervision on the reconstructed expressions during training.
The emotion similarity relies on deep features extracted from a neural network trained for single-image affect (emotion) recognition in-the-wild. 
\model reconstructs 3D face shape on par with  current state-of-the-art methods but outperforms them in terms of the quality of the reconstructed expression. 
Further, using the reconstructed expression parameters for the task of in-the-wild emotion recognition, \model outperforms existing 3DMM-based face reconstruction methods and gives on par results with the best purely image-based method.

In summary, this is the first in-the-wild monocular face reconstruction work that puts explicit emphasis on the \textit{perceptual quality} of the expression and the emotion it communicates instead of standard geometric and photometric losses. 
This presents a new direction for the monocular face reconstruction community. 
This work has potential to further combine the fields of monocular 3D face reconstruction and emotion analysis. 
Further, downstream application of this work can be employed in the industry, including but not limited to gaming, movies, AR/VR and communication.

Of course, any improvement to 3D face acquisition and animation may also enable more realistic `deep fakes.' 
Subtle emotional cues are individualistic and reproducing these could make it harder to detect such fakes.
While cognizant of the risks, we are also sensitive the the importance of facial emotion in human communication.
The trend towards emotional avatars in games and communication is clear. 
If communicative avatars do not properly communicate emotion, that, in itself presents a risk of misunderstandings. 

{
\small
\qheading{Acknowledgement:}
We thank Y. Feng and H. Feng for DECA support and helpful discussions, T. Alexiadis and T. McConnell for help with the perceptual study, S. Zuffi, S. Sanyal, O. Ben-Dov, N. Andreou, P. Patel and P. Forte for proofreading, and A. Toisoul for EmoNet discussions. 
This project has received funding from the European Union’s Horizon 2020 research and innovation programme under the Marie Skłodowska-Curie grant agreement No.860768 
(CLIPE project).

\qheading{Disclosure:}
MJB has received research gift funds from Adobe, Intel, Nvidia, Facebook, and Amazon. 
MJB has financial interests in Amazon, Datagen Technologies, and Meshcapade GmbH.
While TB is part-time employee of Amazon, this research was performed solely at, and funded solely by, MPI. 
}

{\small
\balance

}

\newpage
\nobalance
\appendix
\section{Appendix}
 
\qheading{Discussion of novelty:}
In \supmat we aim to shed more light on the process that eventually led to \model and the challenges that had to be overcome. 
The idea of using deep perceptual losses to supervise face reconstruction is not new. 
A critic might argue, that the novelty of \model is very limited for exactly that reason. 
However, the fact remains that previous SOTA methods have a clear limitation when it comes to reconstructing faces that communicate the correct emotional content.
And from the knowledge of this limitation, we conceived the idea of leveraging emotion recognition, an idea not previously attempted by any work on face reconstruction.
The inventive novelty in \model was in coming up with the idea in the first place. 
This idea, once explained, makes such an intuitive sense, it may lead the reader into thinking it is a straightforward change to an already functioning system.
The idea, although very simple and elegant, was by no means easy to get to work and this is what we aim to explain next.

\qheading{Designing \model:}
Our work starts with the simple idea - how can we employ the findings from emotion recognition to improve face reconstruction?
Leveraging a pretrained SOTA network for emotion recognition, similarly to the way face recognition networks were used seems like a natural choice. 
However, using its final outputs such as the expression class and valence and arousal levels is not sufficient. 
Clearly, these very low-dimensional labels, while they do carry some information about the emotional content, they likely exhibit a lot of ambiguity and are not sufficient to supervise 3D shapes. 
For instance, an expression classified as happy can take on many different shapes (a subtle smile, a big smile with an open mouth, an "inverted" smile, etc.) and similar reasoning could be applied for any other expression and for any levels of valence and arousal as well. 
Hence, these labels most likely do not provide a sufficient supervision signal for geometry. 
The next logical design choice is to leverage high dimensional deep features from a pretrained emotion recognition network.
This choice can only make sense if the emotion feature in question is a "well-behaved" embedding space. 
Ideally we want similar features to represent faces of similar expressions and vice versa. 
Therefore, we conducted an emotion retrieval experiment, using a pretrained publicly available EmoNet model~\cite{toisoul2021estimation} and nearest neighbors search. 
This experiment is discussed in \secref{sec:retrieval}.
Having verified, that similar emotion features retrieve images of geometrically and semantically similar expressions, the next thing to be verified is whether the emotion feature carries a signal that is strong enough, to be utilizable for 3D reconstruction. 
This was particularly challenging and we comment on this further in \secref{sec:optimization}.
Finally, having demonstrated that the emotion recognition features indeed carry enough information in order to supervise the geometry, we can finally incorporate the emotion consistency loss into a face reconstruction framework, arriving at EMOCA. 
In addition to the ablations listed in the main paper, we also add ablations on different architectures and weights for the emotion consistency loss in \secref{sec:consistency}

\section{Implementation details}

\qheading{Emotion recognition metrics:}
In the main paper, we evaluate emotion metrics in the same setting as Toisoul et al.~\cite{toisoul2021estimation}. 
The metrics are defines as follows
$\operatorname{RMSE}$ stands for root mean squared error:
$$
\operatorname{RMSE}(Y, \hat{Y})=\sqrt{\mathbb{E} [(Y-\hat{Y})^{2} ]}.
$$
$\operatorname{SAGR}$ stands for sign agreement and it evaluates whether the predicted value has the same sign as the ground truth:
$$
\operatorname{SAGR}(Y, \hat{Y})=\frac{1}{n} \sum_{i=1}^{n} \delta\left(\operatorname{sign}\left(y_{i}\right), \operatorname{sign}\left(\hat{y}_{i}\right)\right).
$$
Pearson correlation coefficient ($\operatorname{PCC}$) measures the correlation between predictions and GT:
$$
\operatorname{PCC}(Y, \hat{Y})=\frac{\mathbb{E} [(Y-\mu_{Y})(\hat{Y}-\mu_{\hat{Y}})]}{\sigma_{Y} \sigma_{\hat{Y}}}.
$$
Concordance correlation coefficient ($\operatorname{CCC}$) incorporates the $\operatorname{PCC}$ but also penalizes signals which are still correlated according to PCC but have different means:
$$
\operatorname{CCC}(Y, \hat{Y})=\frac{2 \sigma_{Y} \sigma_{\hat{Y}} \operatorname{PCC}(Y, \hat{Y})}{\sigma_{Y}^{2}+\sigma_{\hat{Y}}^{2}+\left(\mu_{Y}-\mu_{\hat{Y}}\right)^{2}}.
$$

\qheading{Emotion recognition loss function:}
We train our emotion networks with the same loss function as defined by Toisoul et al.~\cite{toisoul2021estimation}.
$$
\mathcal{L}_{\text {categories }}(Y, \hat{Y})=\text { Cross entropy }(Y, \hat{Y}) =-\sum_{i=1}^{n} \hat{y}_{i} \log \left(y_{i}\right) \\
$$
The complete loss function for emotion recognition is then defined as:
$$
\begin{aligned}
&\mathcal{L}(Y, \hat{Y})=\mathcal{L}_{\text {categories }}(Y, \hat{Y})+\frac{\alpha}{\alpha+\beta+\gamma} \mathcal{L}_{\mathrm{MSE}}(Y, \hat{Y}) \\
&+\frac{\beta}{\alpha+\beta+\gamma} \mathcal{L}_{\text {PCC }}(Y, \hat{Y})+\frac{\gamma}{\alpha+\beta+\gamma} \mathcal{L}_{\mathrm{CCC}}(Y, \hat{Y}),
\end{aligned}
$$
where $\alpha$, $\beta$ and $\gamma$ are shake-shake regularization coefficients~\cite{shake_gastaldi17} uniformly sampled from the interval $[0,1]$ for each training batch and:
$$
\mathcal{L}_{\mathrm{MSE}}(Y, \hat{Y})=\operatorname{MSE}_{\text {valence }}(Y, \hat{Y})+\operatorname{MSE}_{\text {arousal }}(Y, \hat{Y})
$$
$$
\mathcal{L}_{\mathrm{PCC}}(Y, \hat{Y})=1-\frac{\mathrm{PCC}_{\text {valence }}(Y, \hat{Y})+\mathrm{PCC}_{\text {arousal }}(Y, \hat{Y})}{2}
$$
$$
\mathcal{L}_{\mathrm{CCC}}(Y, \hat{Y})=1-\frac{\mathrm{CCC}_{\text {valence }}(Y, \hat{Y})+\mathrm{CCC}_{\text {arousal }}(Y, \hat{Y})}{2}.
$$
Unlike the work of Toisoul et al.~\cite{toisoul2021estimation}, we do not use knowledge distillation as its improvements are marginal and make the training process much more complex.

\qheading{Image-based emotion recognition:}
We investigate emotion recognition networks based on different architectures, ResNet-50 \cite{He2016_ResNet}, Swin Transformer \cite{Liu2021_SwinTransformer}, and EmoNet \cite{toisoul2021estimation}.
We train all models on AffectNet \cite{affectNet}, using the training/validation/test split proposed by Toisoul et al.~\cite{toisoul2021estimation}.
The ResNet-50 and Swin Transformer based models are pre-trained on ImageNet \cite{Deng2009_ImageNet}.
During training, the training images are sampled such that each of the 7 expression labels appears with the same frequency. 
This sampling is crucial to maximize the performance of the emotion networks, as the AffectNet training set is not balanced. 
We use the Adam optimizer with learning rate of 0.0001, $\beta_1$ = 0.9 and $\beta_2$ = 0.999. 
The batch size used for training is 64. 
Each model is trained for a maximum of 20 epochs with early stopping, and the model with the lowest validation error is selected. 

\qheading{3DMM-based emotion recognition:}

\begin{figure*}[t]
    \offinterlineskip
    \centering
    \includegraphics[width=2.1\columnwidth]{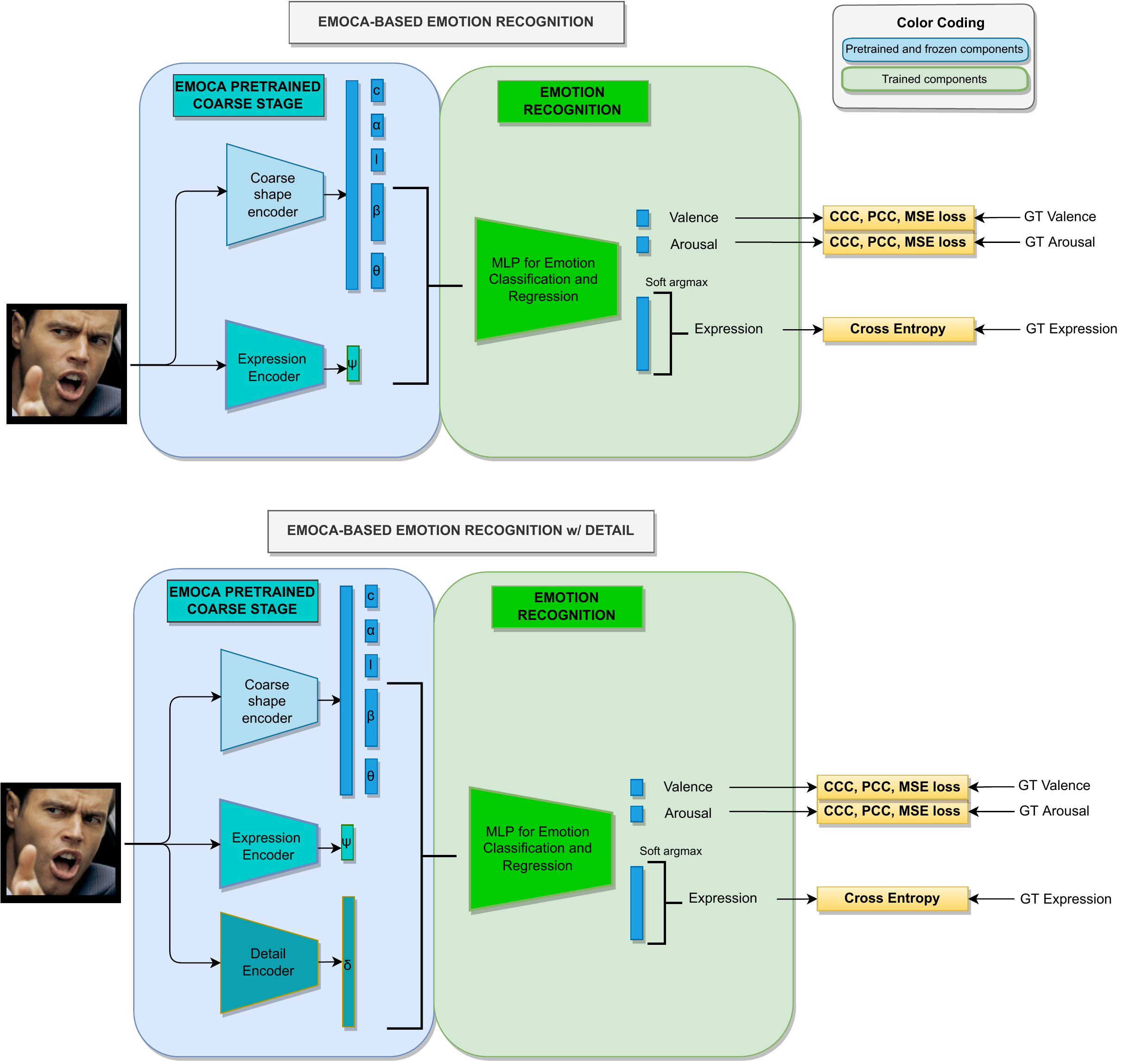}
    \caption{The architecture of EMOCA-based emotion recognition. Top: \model emotion recognition with coarse parameters. From the pretrained coarse stage we extract the shape parameters $\shapecoeff$,  expression parameters $\expcoeff$ and jaw pose $\posecoeff_{jaw}$.  A similar approach is taken for DECA-based recognition, except that DECA does not have a dedicated expression encoder. These are fed to an MLP to regress valence and arousal and classify expression. Bottom: emotion recognition for EMOCA-based reconstruction methods with detail code included.
    }
    \label{fig:recogntion_arch_1}
\end{figure*}

\begin{figure*}[t]
    \offinterlineskip
    \centering
    \includegraphics[width=2.1\columnwidth]{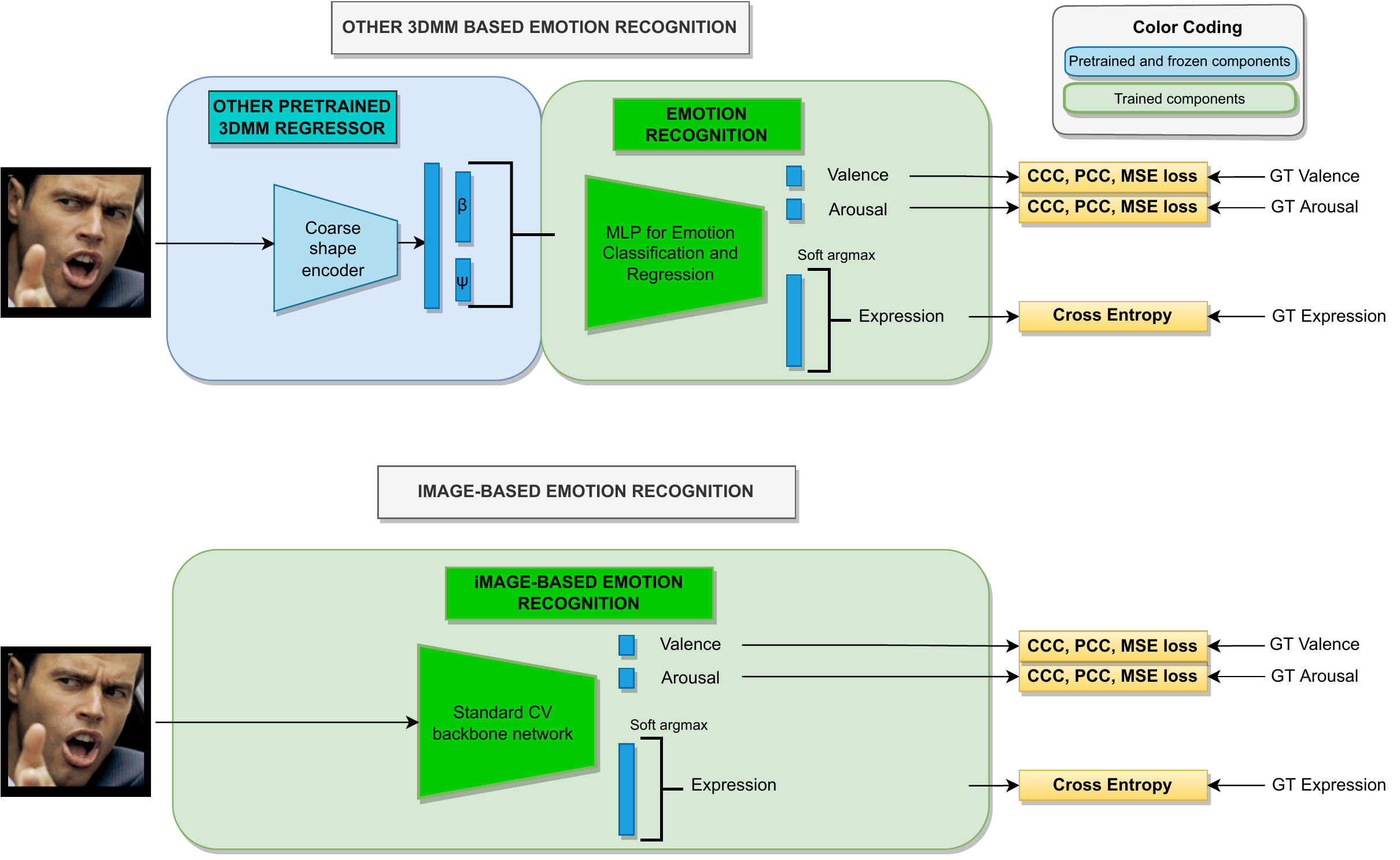}
    \caption{The architecture of other emotion recognition netowrks. Top: emotion recognition for other 3DMM-based reconstruction methods (Deep3DFace \cite{Deng2019}, 3DDFA-V2 \cite{Guo2020towards_3DDFA_V2}, MGCNet \cite{Shang2020_MGCNET}. These have a single decoder that regress to the Basel Face Model \cite{bfm09} parameter space, which does not model jaw pose explicitly. Therefore only $\shapecoeff$ and $\expcoeff$ are considered. Bottom: a standard image-based network trained for emotion recognition.
    Both types of emotion recognition are trained with the same supervision.
    }
    \label{fig:recogntion_arch_2}
\end{figure*}

In Section 5.2 of the paper (Tab.~1) and Table~\ref{table:affewvatest}, we evaluate different face reconstruction methods by recognizing emotions from the regressed 3DMM parameters. 
Specifically, we train a 4-layer MLP with Batch Normalization and LeakyReLUs to output valence and arousal levels and expression classes from the regressed identity and expression parameters (see Figs.~\ref{fig:recogntion_arch_1} and \ref{fig:recogntion_arch_2}  for details).
The size of each hidden layer is 2048. 
We train the 3DMM-based recognition on AffectNet similarly to the image-based emotion recognition. 
The loss function is identical to the one used for image-based emotion recognition. 
The batch size used for training is 64. 
We use the Adam optimizer with a learning rate of 0.0001, $\beta_1=0.9$ and $\beta_2=0.999$.

\qheading{Detail stage training:}
The detail stage training follows the training protocol of DECA \cite{Feng2021_DECA}.
The coarse model part is kept fixed, while detail encoder and decoder are trained. 
This stage uses VGGFace2 \cite{Cao2018_VGGFace2} and VoxCeleb2 \cite{Chung2018_VoxCeleb2} images, due to the necessity of having multiple images per identity. 
We optimize following losses: photometric loss, ID-MRF perceptual loss which encourages reconstruction of higher frequency detail (compared to the coarse mesh), as well as the soft symmetry loss and displacement regularization. 
Further, to disentangle identity and expression dependent details, we employ DECA's detail consistency loss, where each batch contains k images of each subject, and the detail codes are exchanged randomly between the predictions for each identity
For our training, we set k=3 and batch size of 4 identities, totalling 12 input images per batch. 
For more details, see the original DECA publication.

\section{Qualitative evaluation}
In addition to the performance in emotion analysis on the AffectNet dataset in the main paper, we also test \model on AFEW-VA~\cite{AFEW-VA}. The results are reported in \tabref{table:affewvatest}.

\begin{table*}[t]
\centering
\resizebox{0.9\textwidth}{!}{
\begin{tabular}{ll|cccc|cccc}
\toprule
{} &            Model   &  V-PCC $\uparrow$ &  V-CCC $\uparrow$ &  V-RMSE $\downarrow$ &  V-SAGR $\uparrow$ &  A-PCC $\uparrow$ &  A-CCC $\uparrow$&  A-RMSE $\downarrow$ &  A-SAGR $\uparrow$\\
\midrule
  &             EmoNet &   0.59 &   0.54 &    0.22 &    0.61 &   0.55 &   0.49 &    \textit{0.22} &    \textit{0.80} \\ \hline
  &         Deep3DFace &   \textit{0.64} &   \textit{0.59} &    \textit{0.21} &    \textbf{0.65} &   0.55 &   0.48 &    \textbf{0.21} &    \textbf{0.81} \\
  &             ExpNet &   0.31 &   0.25 &    0.27 &    0.55 &   0.36 &   0.30 &    0.24 &    0.79 \\
  &             MGCNet &   0.54 &   0.50 &    0.23 &    0.62 &   0.49 &   0.44 &    0.23 &    0.79 \\
  &              3DDFA &   0.41 &   0.38 &    0.27 &    0.57 &   0.44 &   0.41 &    0.24 &    0.78 \\
  &      DECA (coarse) &   0.57 &   0.53 &    0.23 &    0.62 &   \textit{0.55} &   \textit{0.50} &    \textit{0.22} &    \textbf{0.81} \\
  &      DECA /w detail &   0.57 &   0.53 &    0.23 &    0.63 &   0.53 &   0.49 &    \textit{0.22} &    \textit{0.80} \\ \hline
  &     EMOCA (Ours) &   \textbf{0.65} &   \textbf{0.63} &    \textit{0.21} &    \textit{0.64} &   \textbf{0.57} &   \textbf{0.54} &    \textit{0.22} &    \textit{0.80} \\ 
  &     EMOCA /w detail (Ours) &   \textbf{0.68} &   \textbf{0.65} &    \textbf{0.20} &    \textit{0.64} &   \textit{0.56} &   \textit{0.53} &    \textit{0.22} &    \textit{0.80} \\
\bottomrule
\end{tabular}

}
\caption{
Emotion recognition performance on AFEW-VA~\cite{AFEW-VA}.
All emotion regressors are pretrained on AffectNet and finetuned on the AFEW-VA using 5-fold Cross-Validation (CV). The reported numbers are averaged across the 5-fold CV runs. 
\model performs best, followed by Deep3DFace. Surprisingly, both of these methods outperform EmoNet. Other 3D-based methods follow.
}
\label{table:affewvatest}
\end{table*}

\section{Emotion optimization}
\label{sec:optimization}
We can use our emotion consistency loss for additional tasks.  Here we consider the problem of expression retargeting.
Given two face images, a source identity image $\image_S$ and a target expression image $\image_T$ of potentially two different people with different expressions, poses, cameras, and lighting, our goal is to optimize for the (unknown) target expression $\hat{\expcoeff}_T$. 
Formally, we infer the FLAME parameters $\coarseencoder(\image_S)$ and $\coarseencoder(\image_T)$ for both images.
Then, with some abuse of notation, we render 
$\image_{R}(\expcoeff) = \renderer( \flamev(\shapecoeff_S, \posecoeff_T, \expcoeff), \albedocoeffs_S, \lighting_T, \cam_T )$, the FLAME mesh with source identity shape $\shapecoeff_S$, source albedo $\albedocoeffs_S$, and target pose $\posecoeff_T$, target camera $\cam_T$, target lighting $\lighting_T$, and the optimization expression parameters $\expcoeff$.
We then extract the emotion features of the rendering $\emovec_R(\expcoeff) = \emonet(\image_R(\expcoeff))$ and the target image $\emovec_T = \emonet(\image_T)$, and optimize:
\begin{equation}
    \hat{\expcoeff}_T = \argmin_{\expcoeff} \emometric(  \emovec_R(\expcoeff), \emovec_T) + \lambda_{\expcoeff} L_{\expcoeff},
\end{equation}
with $\emometric(\emovec_1, \emovec_2) = \norm{\emovec_1  - \emovec_2}_2$, expression regularizer $L_{\expcoeff} = \norm{\expcoeff}_2^2$, and regularizer weight $\lambda_{\expcoeff} = 1e-3$.
We use gradient descent for the optimization.
Below we show optimization results, and an analysis of the convergence and sensitivity to the initialization.

\begin{figure}[t]
    \offinterlineskip
    \centering
    \includegraphics[width=0.175\columnwidth]{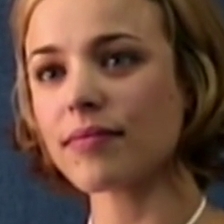} 
    \includegraphics[width=0.175\columnwidth]{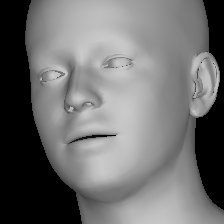} 
    \includegraphics[width=0.175\columnwidth]{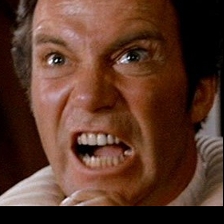} 
    \includegraphics[width=0.175\columnwidth]{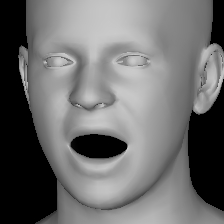} 
    \includegraphics[width=0.175\columnwidth]{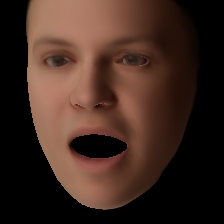} \\ 
    
    \includegraphics[width=0.175\columnwidth]{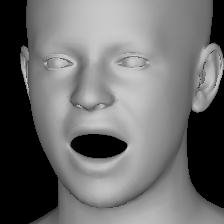}
    \includegraphics[width=0.175\columnwidth]{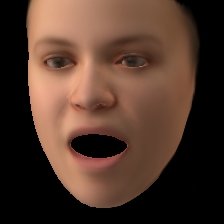}
    \includegraphics[width=0.175\columnwidth]{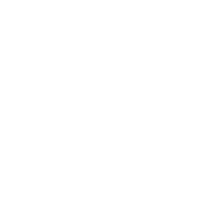}
    \includegraphics[width=0.175\columnwidth]{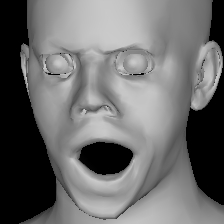}
    \includegraphics[width=0.175\columnwidth]{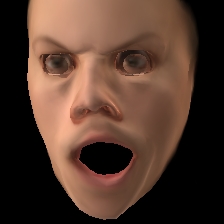} \\
    
    \includegraphics[width=0.175\columnwidth]{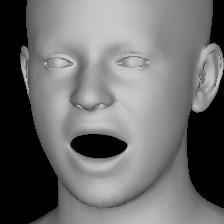}
    \includegraphics[width=0.175\columnwidth]{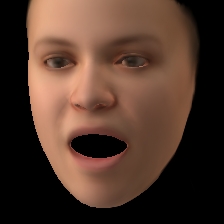}
    \includegraphics[width=0.175\columnwidth]{fig/retargetting/white.jpg}
    \includegraphics[width=0.175\columnwidth]{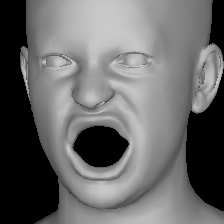}
    \includegraphics[width=0.175\columnwidth]{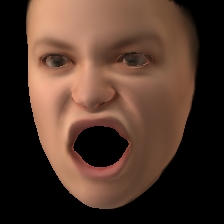} \\ 
    
    \includegraphics[width=0.175\columnwidth]{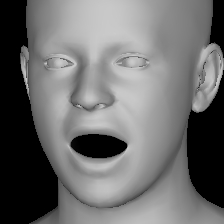}
    \includegraphics[width=0.175\columnwidth]{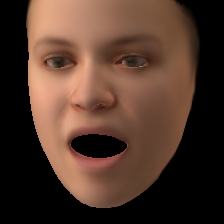}
    \includegraphics[width=0.175\columnwidth]{fig/retargetting/white.jpg}
    \includegraphics[width=0.175\columnwidth]{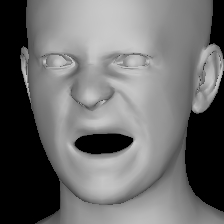}
    \includegraphics[width=0.175\columnwidth]{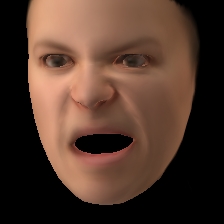}
    
    \caption{Emotion optimization example. 
    The first row contains a source image, its DECA reconstruction, a target image, its DECA reconstruction, and the colored reconstruction. 
    The following rows contain: the initialization of the optimization w/o and w/ color (left) and the optimization result w/o and w/ color (right). 
    The different rows use different emotion recognition networks for optimization. 
    The second row uses the original released EmoNet, the third row a self-trained EmoNet, and the bottom row using our ResNet-50 model. 
    While EmoNet gives SOTA emotion recognition results, it is less suitable for our task of emotion-driven expression optimization or reconstruction. 
    }
    \label{fig:opt_arch_comp}
\end{figure}

\qheading{On Emotion Network Architecture:}
Figure~\ref{fig:opt_arch_comp} shows emotion optimization results using different emotion recognition network. 
This indicates that the original released EmoNet is not suitable for emotion optimization. 
Instead, we use the ResNet-50 architecture as default model. 

\begin{figure}[t]
    \offinterlineskip
    \centering
    \includegraphics[width=0.175\columnwidth]{fig/opt/source.jpg} 
    \includegraphics[width=0.175\columnwidth]{fig/opt/source_shape.jpg} 
    \includegraphics[width=0.175\columnwidth]{fig/opt/resnet/0267/0000_0267_00.jpg} 
    \includegraphics[width=0.175\columnwidth]{fig/opt/resnet/0267/exp_pose_jaw_from_target/target/coarse_shape.jpg} 
    \includegraphics[width=0.175\columnwidth]{fig/opt/resnet/0267/exp_pose_jaw_from_target/target/coarse_output.jpg} \\ 
    
    \includegraphics[width=0.175\columnwidth]{fig/opt/resnet/0267/exp_pose_jaw_from_target/init/coarse_shape.jpg}
    \includegraphics[width=0.175\columnwidth]{fig/opt/resnet/0267/exp_pose_jaw_from_target/init/coarse_output.jpg}
    \includegraphics[width=0.175\columnwidth]{fig/retargetting/white.jpg}
    \includegraphics[width=0.175\columnwidth]{fig/opt/resnet/0267/exp_pose_jaw_from_target/best/coarse_shape.jpg}
    \includegraphics[width=0.175\columnwidth]{fig/opt/resnet/0267/exp_pose_jaw_from_target/best/coarse_output.jpg} \\
    
    \includegraphics[width=0.175\columnwidth]{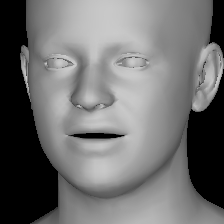}
    \includegraphics[width=0.175\columnwidth]{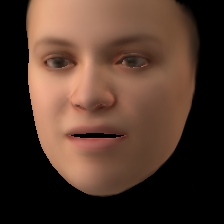}
    \includegraphics[width=0.175\columnwidth]{fig/retargetting/white.jpg}
    \includegraphics[width=0.175\columnwidth]{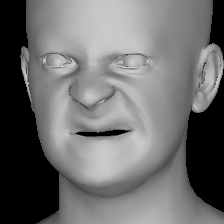}
    \includegraphics[width=0.175\columnwidth]{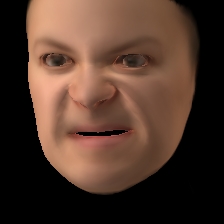} \\ 
    
    \includegraphics[width=0.175\columnwidth]{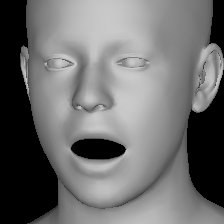}
    \includegraphics[width=0.175\columnwidth]{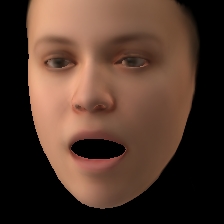}
    \includegraphics[width=0.175\columnwidth]{fig/retargetting/white.jpg}
    \includegraphics[width=0.175\columnwidth]{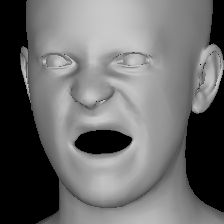}
    \includegraphics[width=0.175\columnwidth]{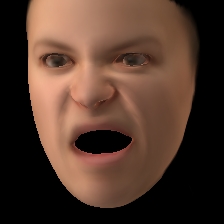} \\
    
    \includegraphics[width=0.175\columnwidth]{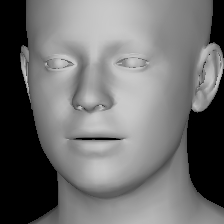}
    \includegraphics[width=0.175\columnwidth]{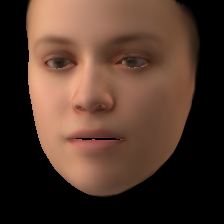}
    \includegraphics[width=0.175\columnwidth]{fig/retargetting/white.jpg}
    \includegraphics[width=0.175\columnwidth]{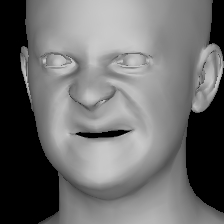}
    \includegraphics[width=0.175\columnwidth]{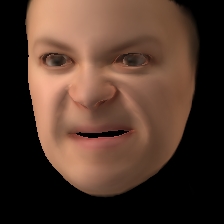} \\
    
    \includegraphics[width=0.175\columnwidth]{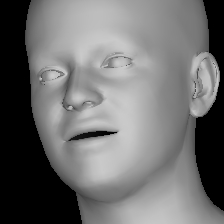}
    \includegraphics[width=0.175\columnwidth]{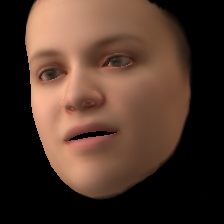}
    \includegraphics[width=0.175\columnwidth]{fig/retargetting/white.jpg}
    \includegraphics[width=0.175\columnwidth]{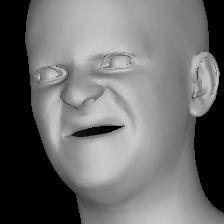}
    \includegraphics[width=0.175\columnwidth]{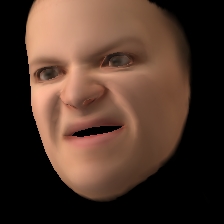} \\
    
    \includegraphics[width=0.175\columnwidth]{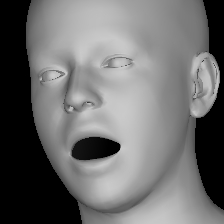}
    \includegraphics[width=0.175\columnwidth]{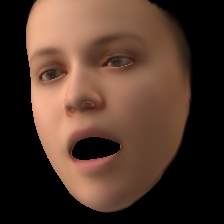}
    \includegraphics[width=0.175\columnwidth]{fig/retargetting/white.jpg}
    \includegraphics[width=0.175\columnwidth]{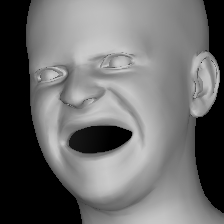}
    \includegraphics[width=0.175\columnwidth]{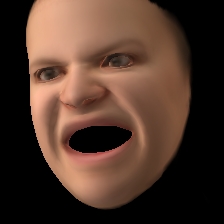} \\
    
    \includegraphics[width=0.175\columnwidth]{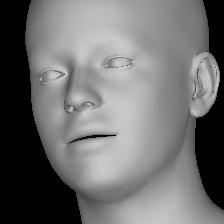}
    \includegraphics[width=0.175\columnwidth]{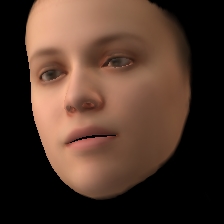}
    \includegraphics[width=0.175\columnwidth]{fig/retargetting/white.jpg}
    \includegraphics[width=0.175\columnwidth]{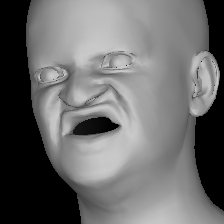}
    \includegraphics[width=0.175\columnwidth]{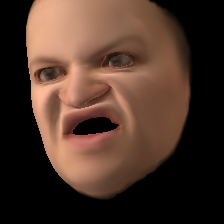} \\
    
    \caption{Sensitivity of the emotion optimization to initialization. 
     The first row contains a source image, its DECA reconstruction, a target image, its DECA reconstruction, and the colored reconstruction. 
    The following rows contain: the initialization of the optimization w/o and w/ color (left) and the optimization result w/o and w/ color (right).     
    Note that the optimization process is only modifying the expression coefficients $\expcoeff$ and not the jaw rotation $\posecoeff_{jaw}$. 
    While the process usually converges to meaningful results, the most favorable outcome is obtained, when initializing the process with the target expression coefficients $\expcoeff$ and pose $\posecoeff$, which correspond to the second row.  
    }
    \label{fig:opt_init_comp}
\end{figure}

\qheading{On Initialization:}
Figure~\ref{fig:opt_init_comp} further shows the influence of the initialization on the optimized emotion. 
These results demonstrate that 3DMMs, when rendered, can in fact be animated with a deep perceptual emotion similarity loss.

\qheading{On Jaw Optimization:}
A perceptive reader may ask, why we optimize only for the expression parameters $ \expcoeff $ and not also for the jaw pose $ \posecoeff_{jaw} $. 
After all, the jaw position most certainly has an effect on the perceived emotion. 
We have struggled with the jaw optimization issue for quite a long time, unable to get acceptable results as the jaw pose parameter optimization makes this optimization unstable - the jaw would always be posed to an unrealistic or at least very incorrect pose. Fixing the jaw pose to a reasonable estimate however (such as DECA's prediction) makes the optimization stable and produces good results. We hypothesise that this instability could be caused by the following: 
\begin{enumerate}
    \item FLAME is missing a comprehensive prior for the jaw pose. We experimented with simplistic hand-crafted priors (such as distance or squared distance from the expected pose) but this did not yield any improvement. It is possible that the creating a more comprehensive prior (other than the Gaussian prior for FLAME's expression space), a prior that entangles the expression and jaw pose spaces is necessary. This makes for an interesting direction for future work. 
    \item Emotion optimization involves optimizing a deep feature vector and while we have demonstrated that similar emotion features belong to similar expressions, we have not eliminated the possibility, that the emotion network can be ``attacked'' to produce the desired features with a distorted images. An optimization process, in which the jaw is not fixed could results in an adversarial attack on the network that forces it to produce a similar emotion feature vector. 
\end{enumerate}

\section{Perceptual study}

\begin{figure*}[t]
    \offinterlineskip
    \centering
    \includegraphics[width=2.2\columnwidth]{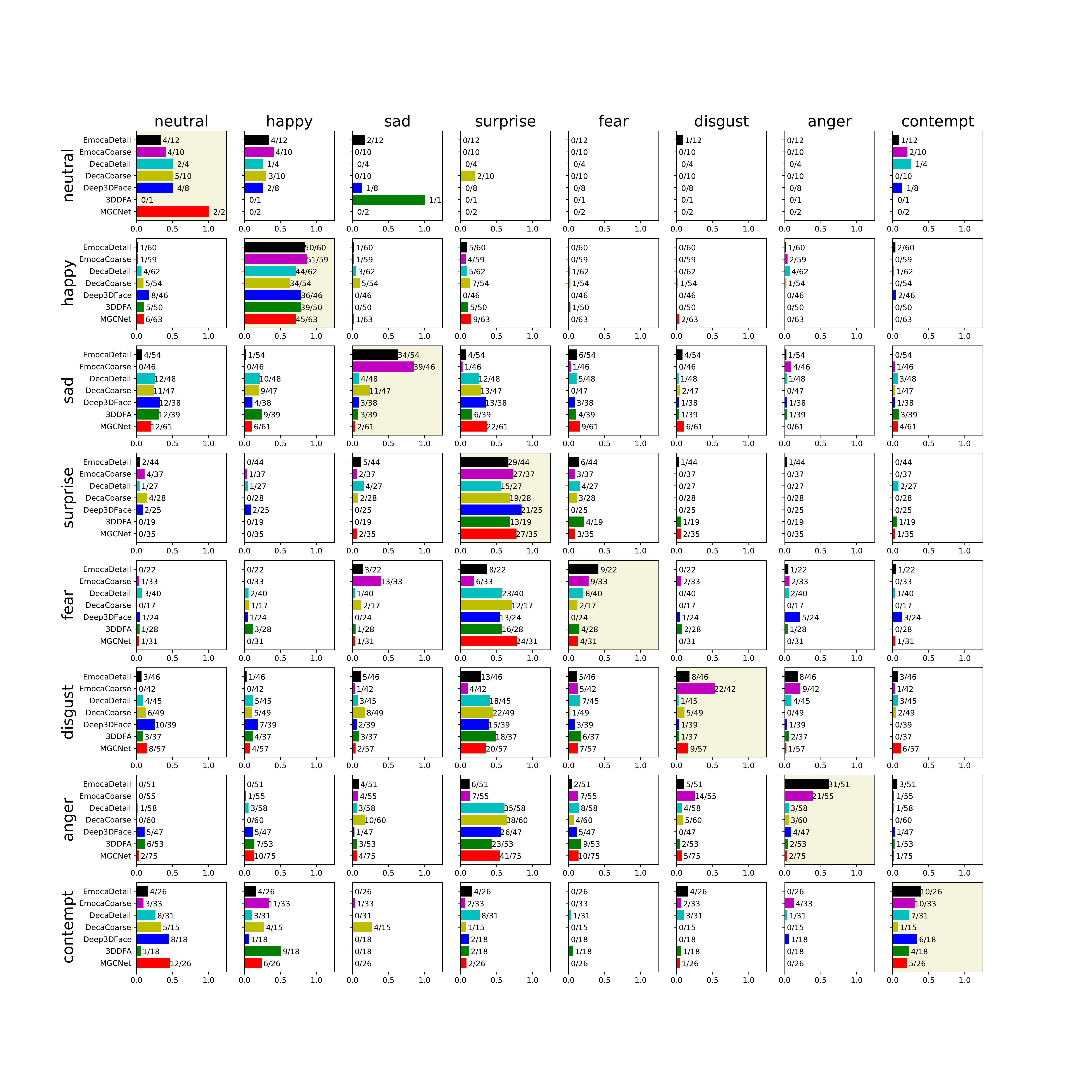} 
    \caption{ This figure contains the confusion matrices of participant's labels of the real image and the reconstructed images for each method. 
    The x-axis of each cell gives the ratio of participants' reconstruction labels and real image labels and the absolute number is written next to each bar. 
    The accuracy of each method for a particular expression class is on the diagonal. You can see that both variants of \model (detail and coarse) are superior to the other methods. Furthermore, off-diagonal you can observe how the label of meshes reconstructed by \model is much less confused for other labels, compared to other methods. Finally,  the confusion matrix highlights how other methods are not capable of producing expressions of fear, disgust and anger. Instead these are confused with surprise. \model does not suffer from the same limitation. However, participants did have some trouble distinguishing reconstructions of disgust and anger. Please note that the first row (neutral) shows a small number of samples. This happens because our perceptual study did not contain neutral images.
    }
    \label{fig:preceptual_user_gt}
\end{figure*}

\begin{figure*}[t]
    \offinterlineskip
    \centering
    \includegraphics[width=2.2\columnwidth]{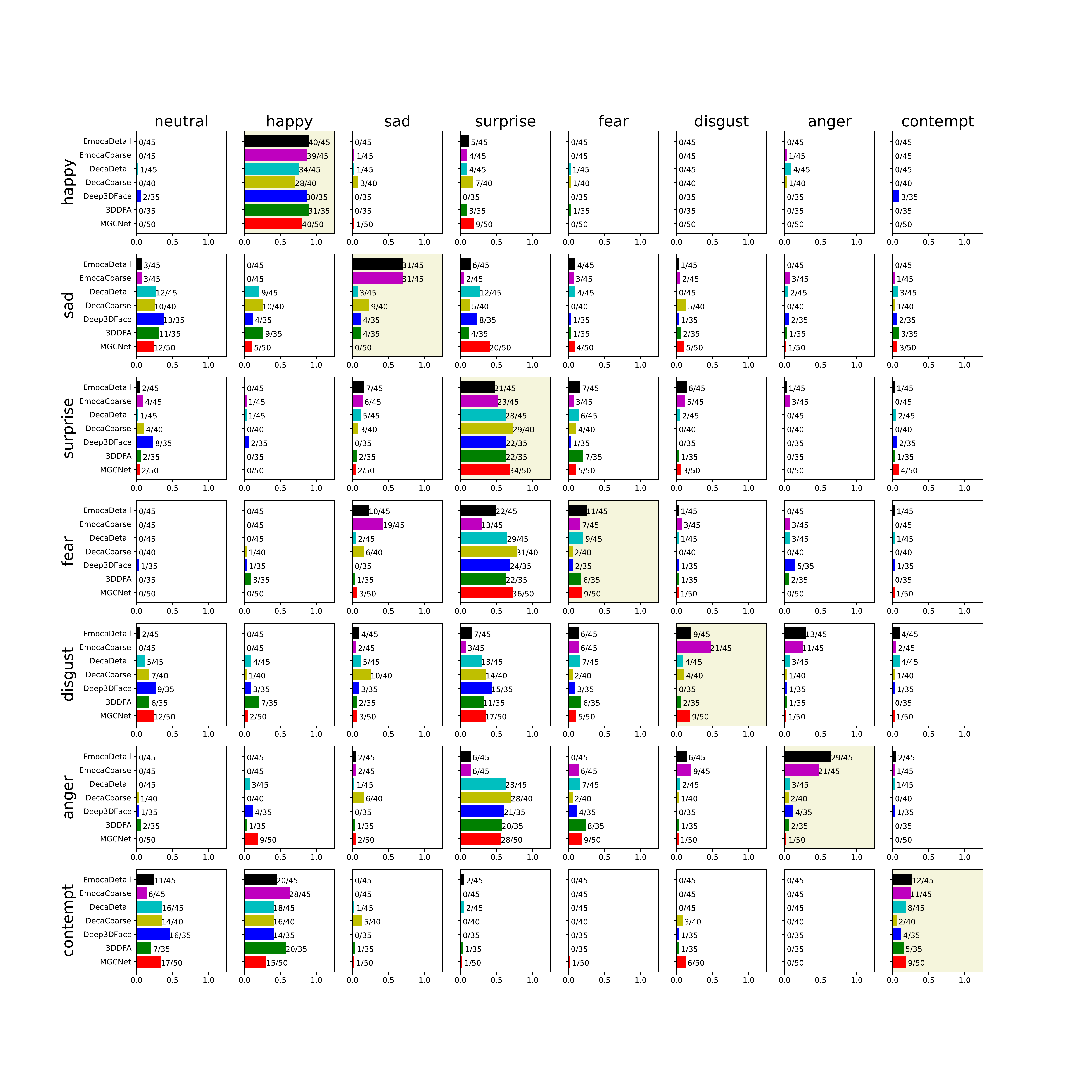} 
    \caption{ This figure contains the confusion matrices of participant's labels of the reconstructions w.r.t. to the ground truth labels (as opposed to users' subjective labels, which you can find in \figref{fig:preceptual_user_gt}. Please note that neutral expressions were not given in the study, which is why the matrix only has six rows (neutral excluded).
    }
    \label{fig:preceptual_gt}
\end{figure*}

\begin{figure*}[t]
    \offinterlineskip
    \centering
    \includegraphics[width=2.2\columnwidth]{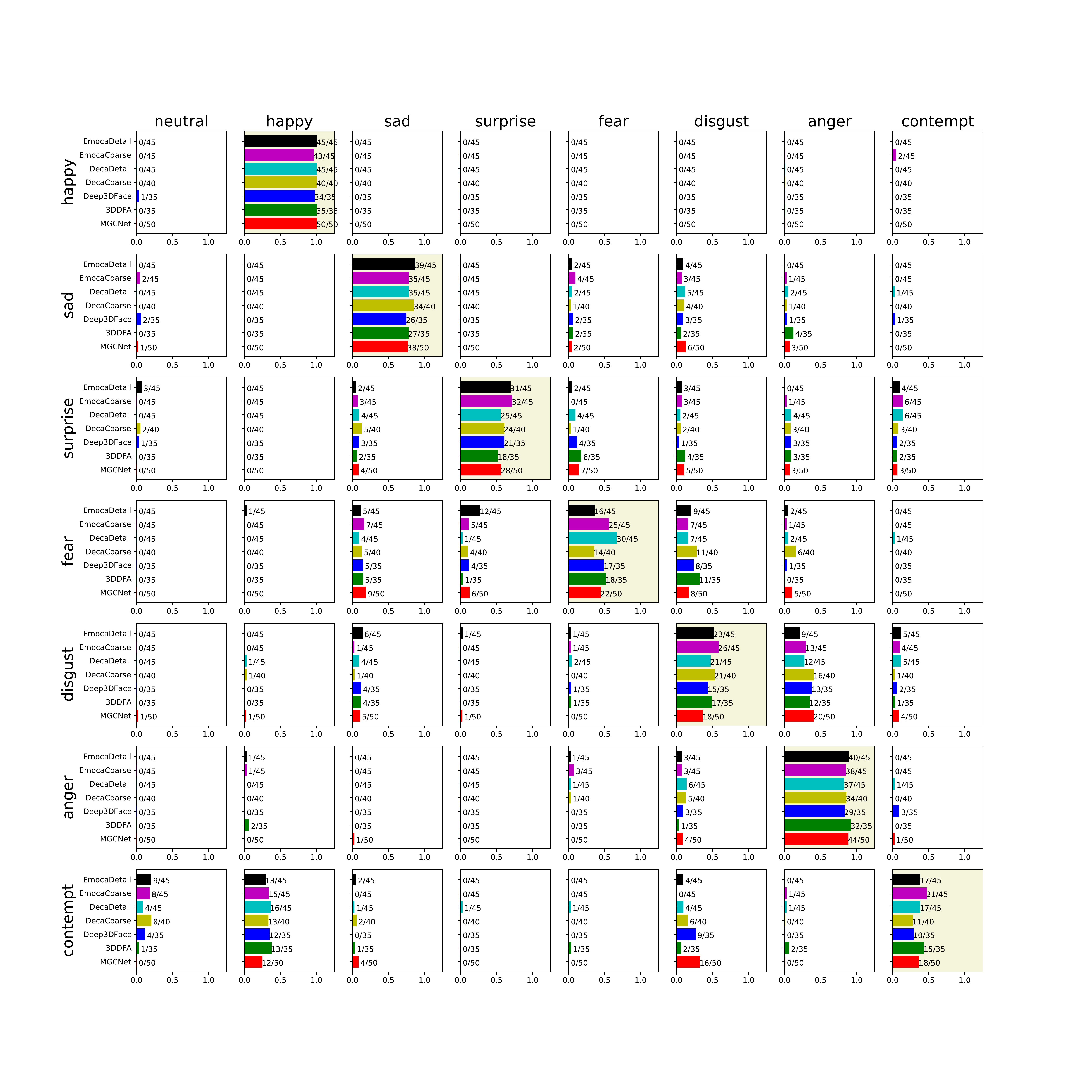} 
    \caption{ This figure contains the confusion matrices of participant's labels of the real images w.r.t. to ground truth images. While this figure does not compare the performance of methods, it serves as a baseline comparison to \figref{fig:preceptual_gt}. Classifying expression is subjective. While our participants mostly agreed with our ground truth, there were disagreements for the negatively charged expressions of fear, disgust, anger and particularly contempt.
    }
    \label{fig:preceptual_im}
\end{figure*}

Section 5.2 of the paper evaluates the amount emotion conveyed by the reconstructed 3D geometry in a perceptual study. 
Figure \ref{fig:preceptual_user_gt} gives the full confusion matrix of the participants' labels of real images (rows) and the labels of the reconstructions (columns).
Figure~\ref{fig:preceptual_gt} further compares the ground truth emotion labels with the participants' classifications of the reconstructions. 
For completeness, we also include the confusion matrix of participants' labeling of the real images in \figref{fig:preceptual_im}.

\section{Emotion consistency}
\label{sec:consistency}

\begin{figure*}[t]
    \offinterlineskip
    \centering
    \includegraphics[width=0.18\columnwidth]{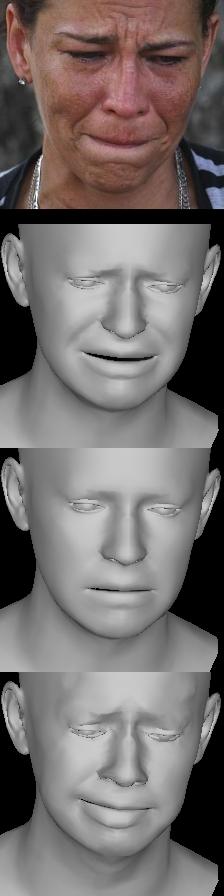}
    \includegraphics[width=0.18\columnwidth]{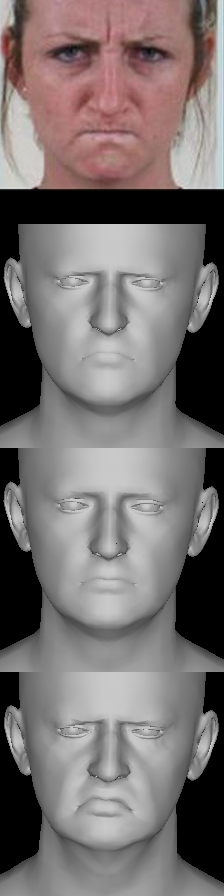} 
    \includegraphics[width=0.18\columnwidth]{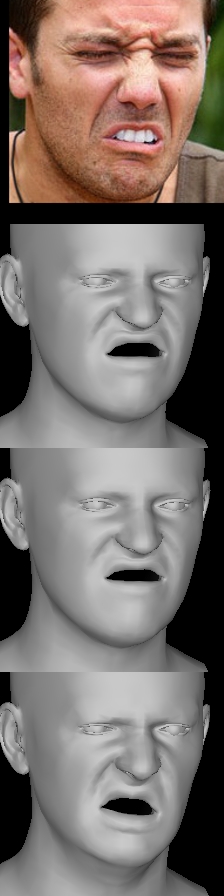} 
    \includegraphics[width=0.18\columnwidth]{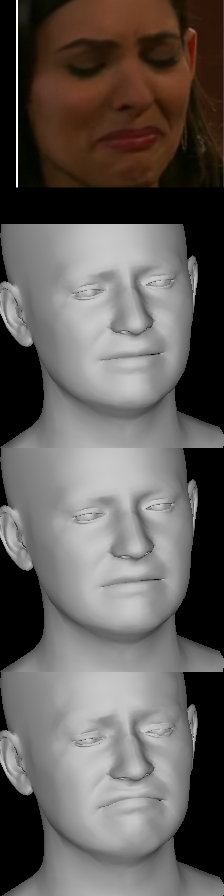} 
    \includegraphics[width=0.18\columnwidth]{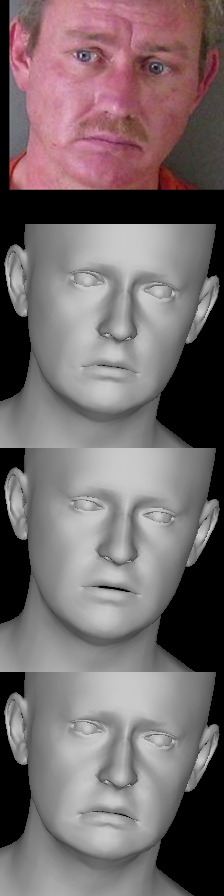} 
    \includegraphics[width=0.18\columnwidth]{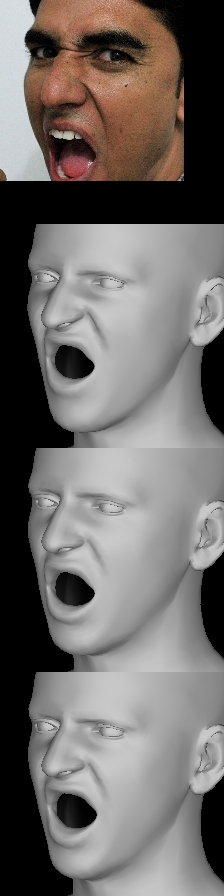} 
    \includegraphics[width=0.18\columnwidth]{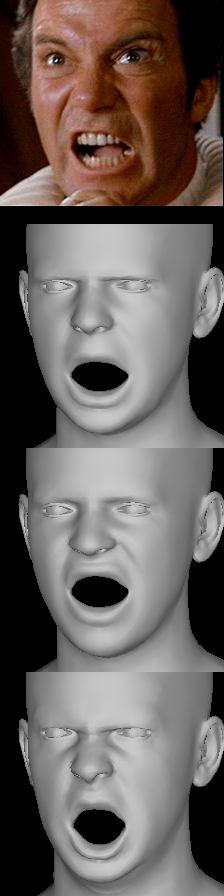} 
    \includegraphics[width=0.18\columnwidth]{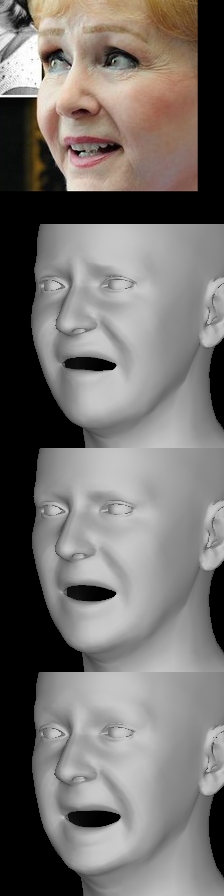} 
    \includegraphics[width=0.18\columnwidth]{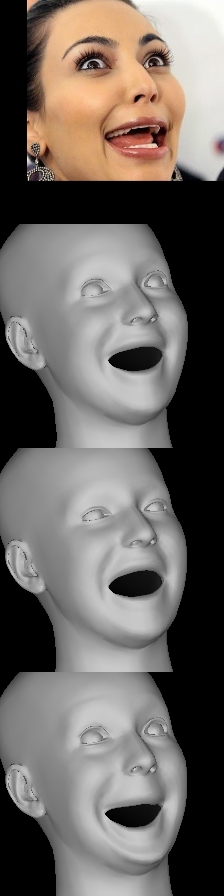} 
    \includegraphics[width=0.18\columnwidth]{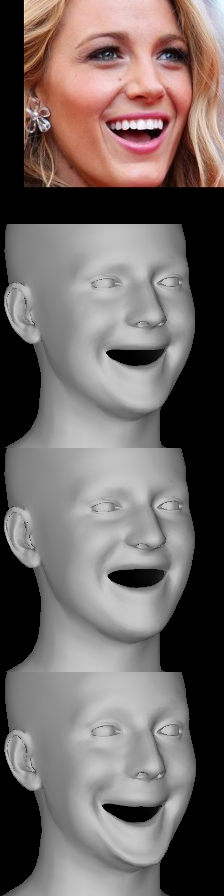} 
    \includegraphics[width=0.18\columnwidth]{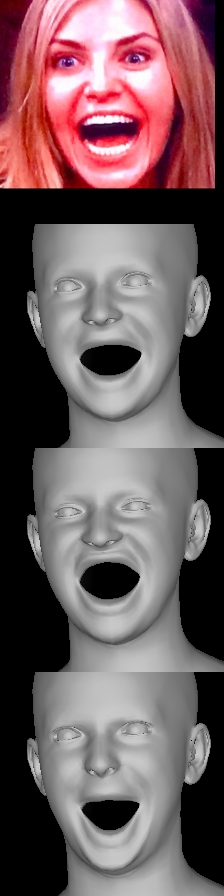} 

    \caption{Comparison of different \model models, supervised by different emotion networks. From top to bottom: ResNet-50 \cite{He2016_ResNet}, SWIN-B \cite{Liu2021_SwinTransformer}, EmoNet \cite{toisoul2021estimation}. All three networks affect the reconstruction in different ways. \model-ResNet produces the best visual results and is our model of choice. \model-SWIN produces results of slightly lower visual quality. Finally, \model-EmoNet sometimes produces unrealistic expressions, which makes EmoNet less suitable for this task.
    }
    \label{fig:emo_arch}
\end{figure*}

\qheading{Emotion network architecture:}
The choice of architecture for emotion supervision plays a critical role. 
While all architectures perform comparatively well on the emotion recognition task, they are not equally suitable as supervision for our 3D face reconstruction task. 
\figref{fig:emo_arch} visually compares \model models trained with different emotion recognition networks as supervision. 
Again, the SOTA emotion recognition architecture - EmoNet, is not suitable as it produces unacceptable artifacts. Furthermore, the SWIN~\cite{Liu2021_SwinTransformer} transformer backbone, which is considered to be superior to the ResNet~\cite{He2016_ResNet} architecture, also produces some undesirable artifacts. Hence, the ResNet backbone was used for the final model of the emotion recognition network.

\begin{figure*}[t]
    \offinterlineskip
    \includegraphics[width=0.18\columnwidth]{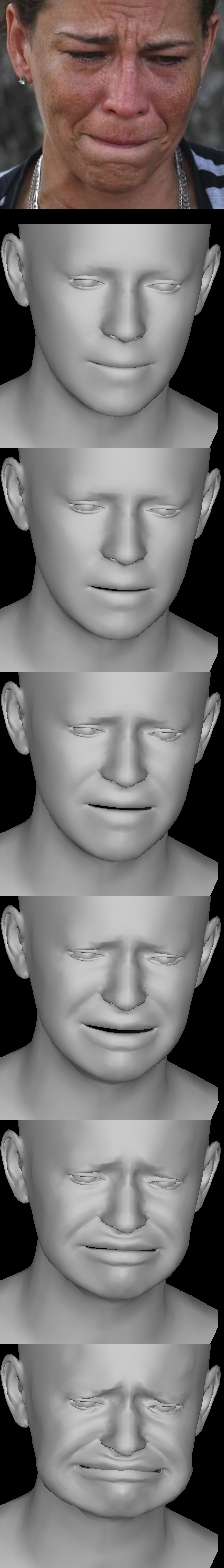}
    \includegraphics[width=0.18\columnwidth]{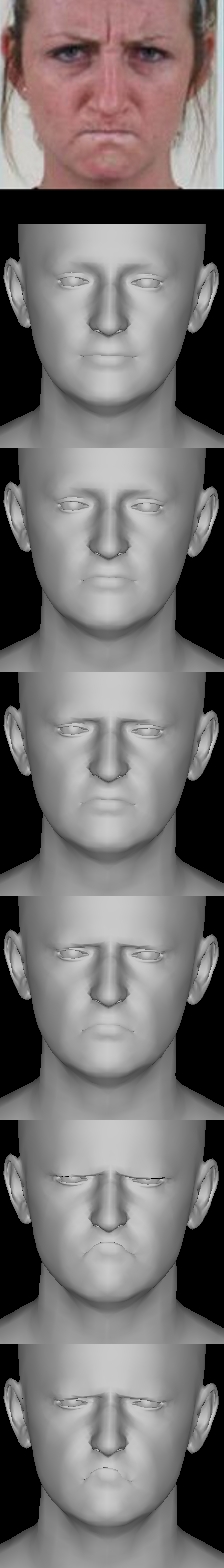} 
    \includegraphics[width=0.18\columnwidth]{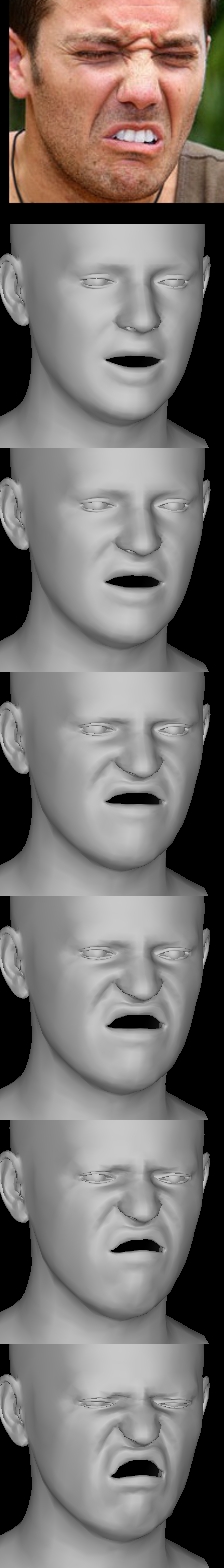} 
    \includegraphics[width=0.18\columnwidth]{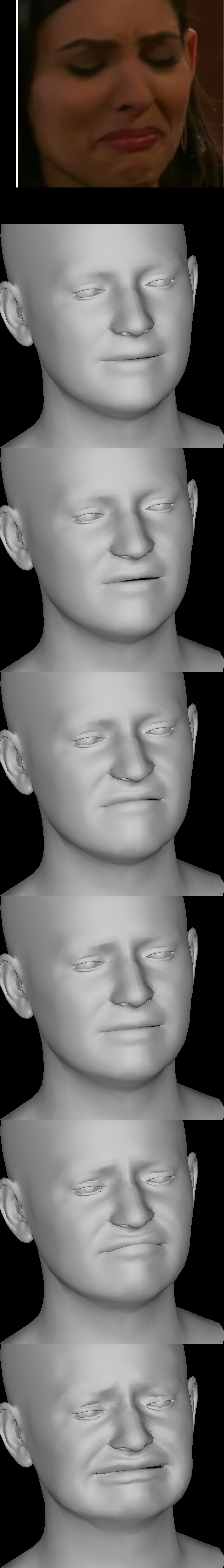} 
    \includegraphics[width=0.18\columnwidth]{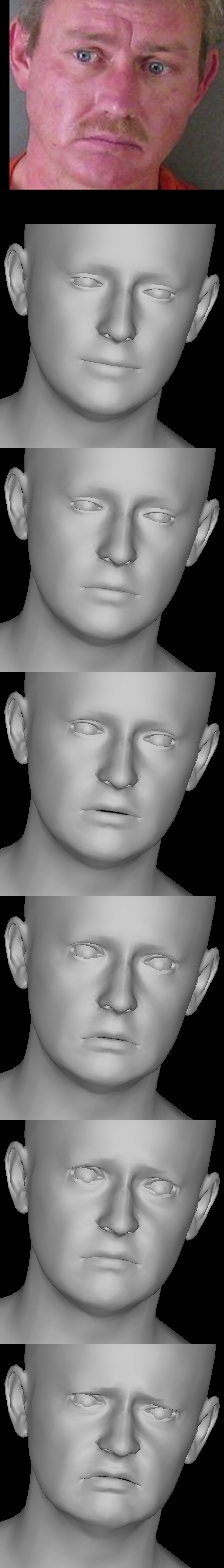} 
    \includegraphics[width=0.18\columnwidth]{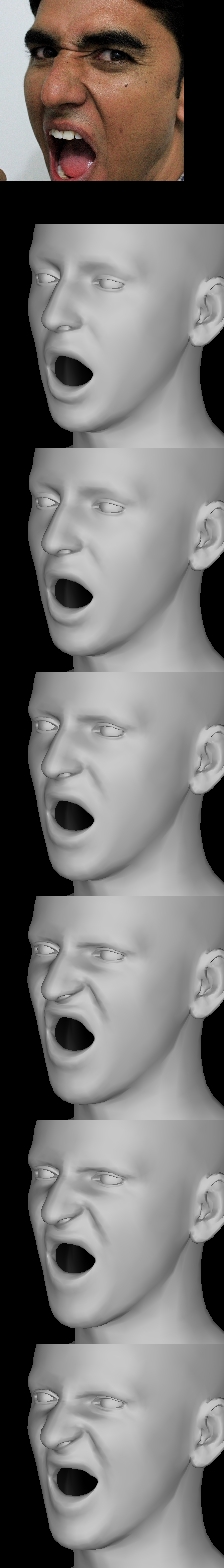} 
    \includegraphics[width=0.18\columnwidth]{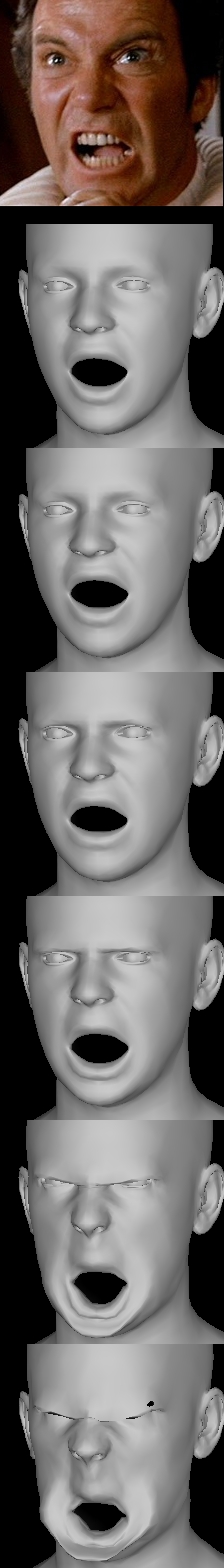} 
    \includegraphics[width=0.18\columnwidth]{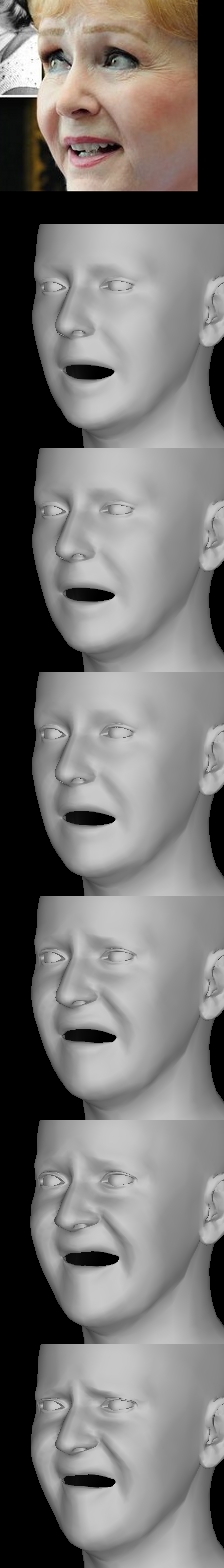} 
    \includegraphics[width=0.18\columnwidth]{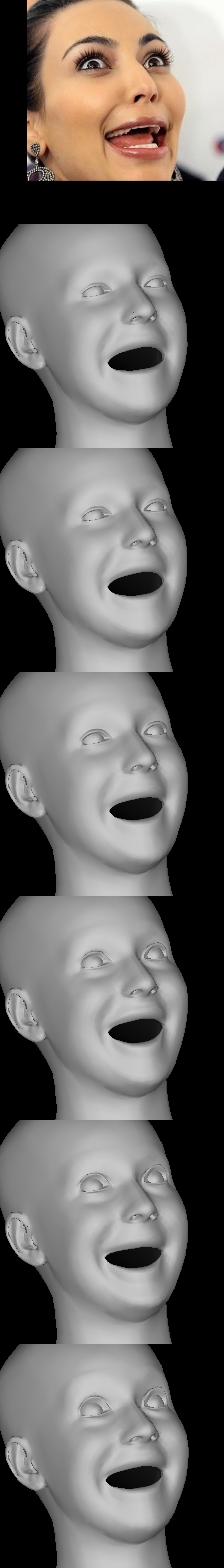} 
    \includegraphics[width=0.18\columnwidth]{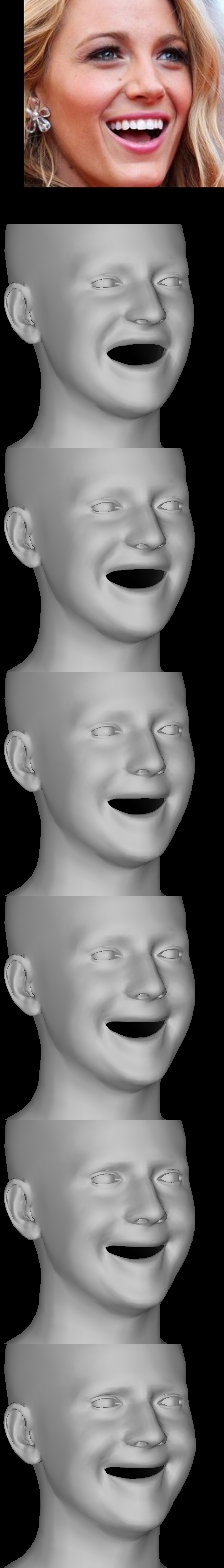} 
    \includegraphics[width=0.18\columnwidth]{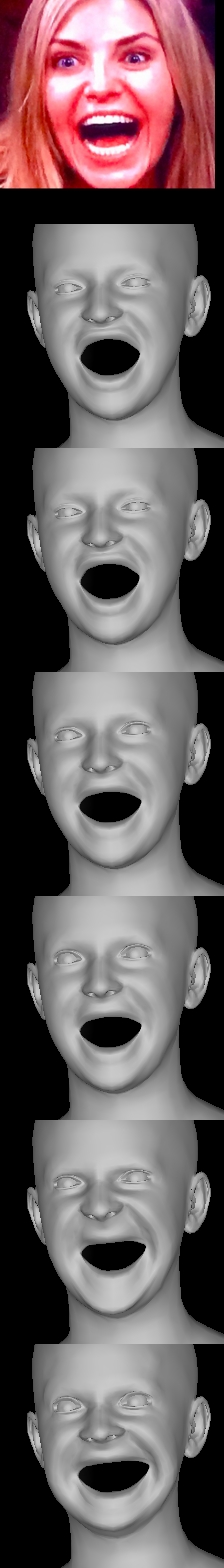} 
    \caption{Comparison of models trained with different weights of the emotion consistency loss $\lambda_{emo}$. The emotion network used was ResNet-50 \cite{He2016_ResNet}. Top row consists of input images. Different values of $\lambda_{emo}$ follow. From top to bottom 0, 0.1, 0.5, 1 (final \model), 5, 10.
    }
    \label{fig:emoresnet_weight}
\end{figure*}

\begin{figure*}[t]
    \offinterlineskip
    \centering
    \includegraphics[width=0.18\columnwidth]{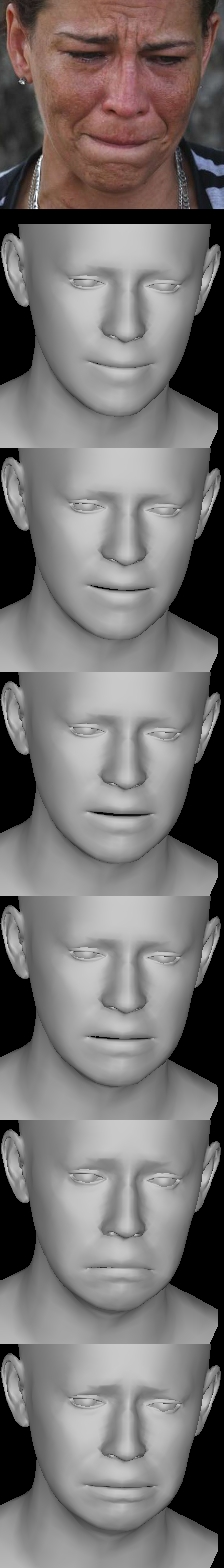}
    \includegraphics[width=0.18\columnwidth]{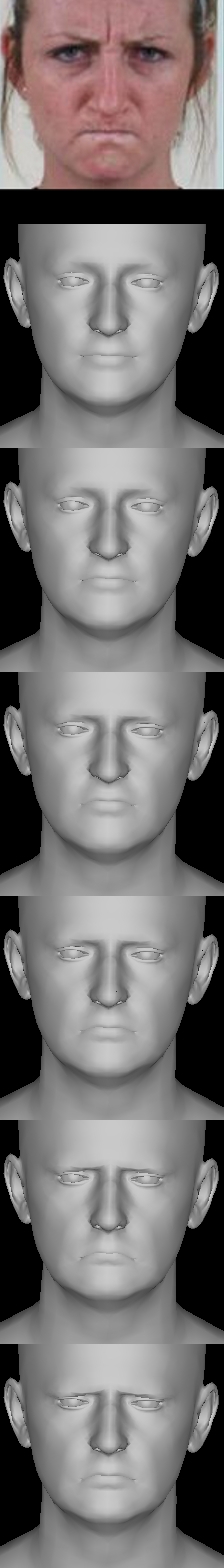} 
    \includegraphics[width=0.18\columnwidth]{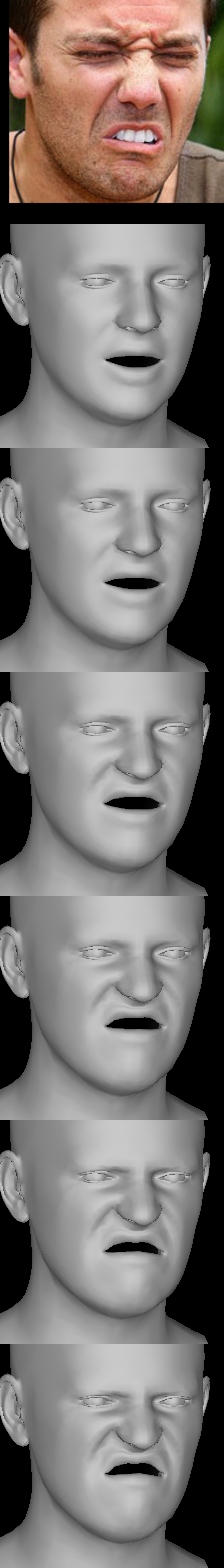} 
    \includegraphics[width=0.18\columnwidth]{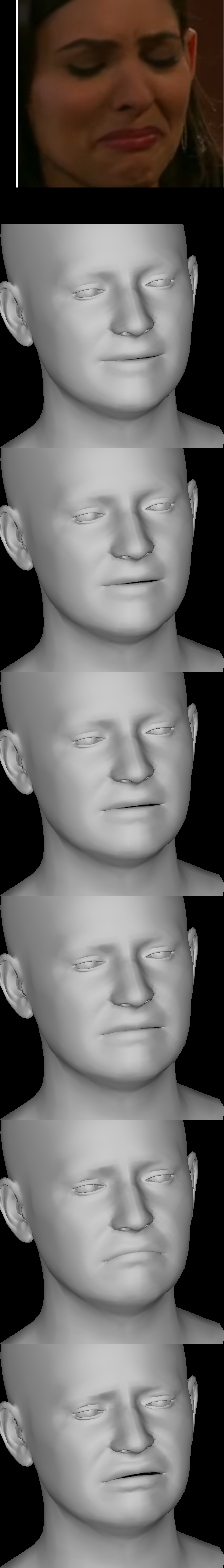} 
    \includegraphics[width=0.18\columnwidth]{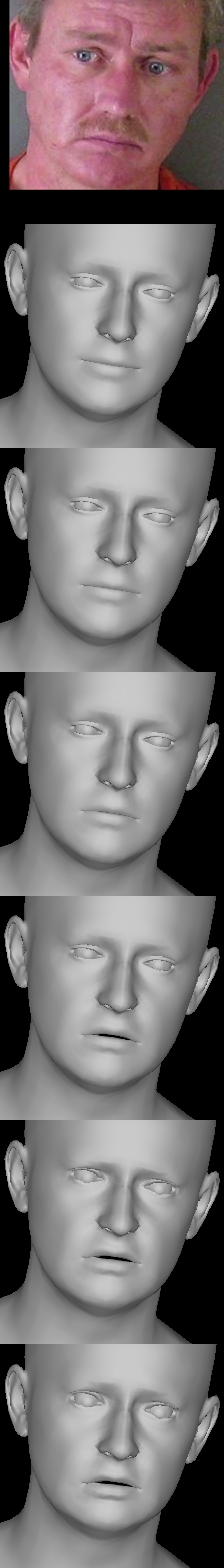} 
    \includegraphics[width=0.18\columnwidth]{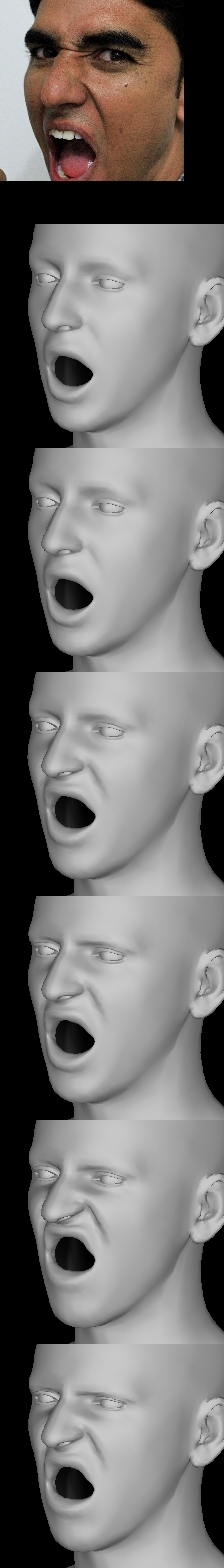} 
    \includegraphics[width=0.18\columnwidth]{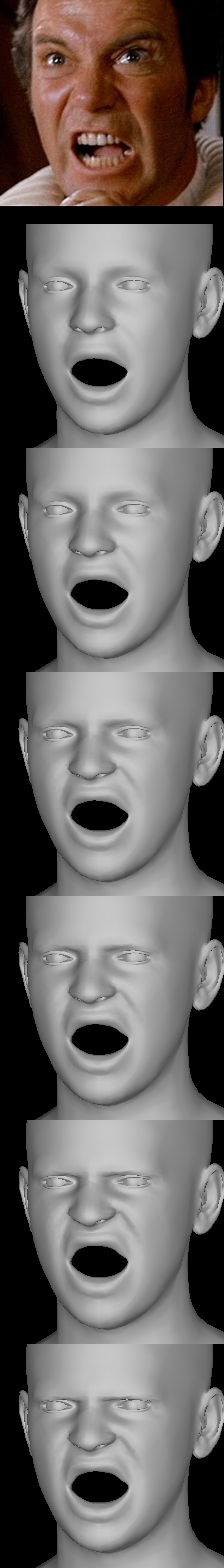} 
    \includegraphics[width=0.18\columnwidth]{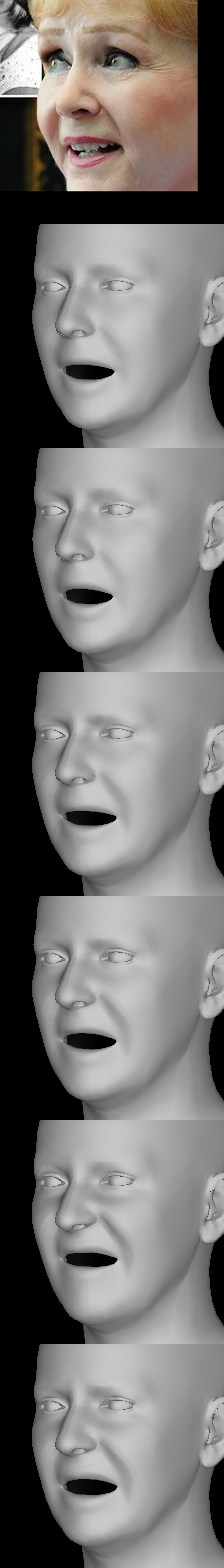} 
    \includegraphics[width=0.18\columnwidth]{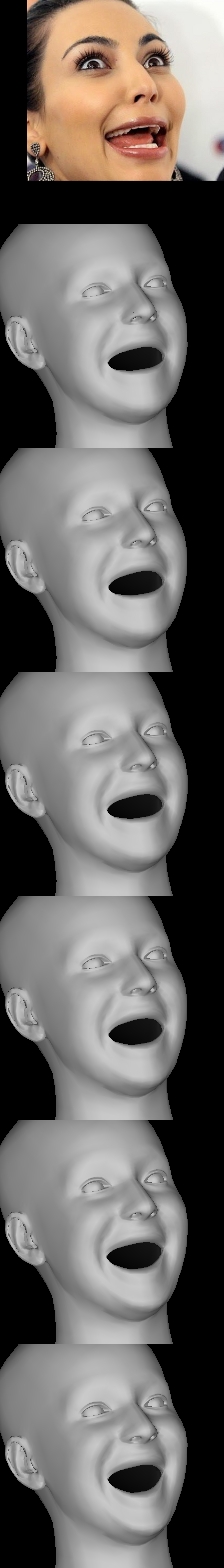} 
    \includegraphics[width=0.18\columnwidth]{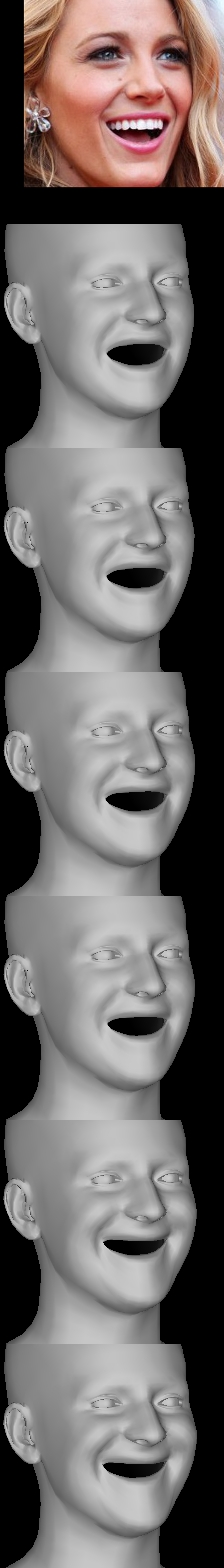} 
    \includegraphics[width=0.18\columnwidth]{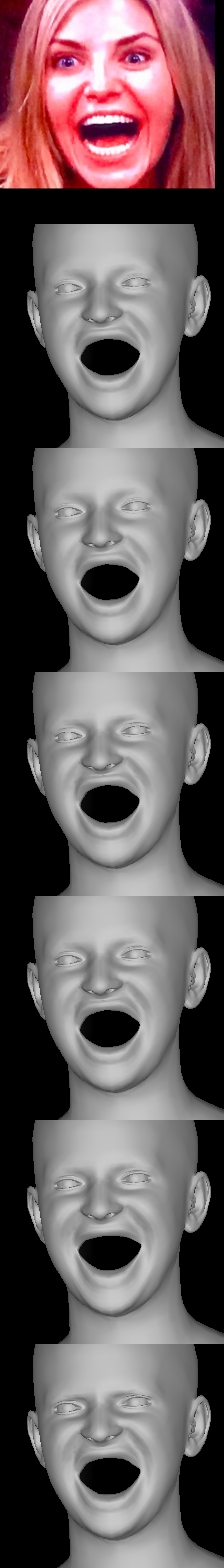} 

    \caption{Comparison of models trained with different weights of the emotion consistency loss $\lambda_{emo}$. The emotion network used was SWIN-B \cite{Liu2021_SwinTransformer}. Top row consists of input images. Different values of $\lambda_{emo}$ follow. From top to bottom 0, 0.1, 0.5, 1 , 5, 10. While SWIN-B suffers from fewer artifacts compared to ResNet-50 when changing the weight, we have deemed the visual quality of results produce by a ResNet-supervised \model slightly better, which is why ResNet was selected for the final model. 
    }
    \label{fig:emoswin_weight}
\end{figure*}

\qheading{Emotion consistency weight:}
We have experimented with different values of the emotion consistency loss weight term $\lambda_{emo}$. This is a crucial factor of successfully training \model. If the weight is too small, the emotion is not captured well enough. At the same time, high values lead to unnaturally over-exaggerated expressions. A visual ablation of this phenomenon can be found in \figref{fig:emoresnet_weight} and \figref{fig:emoswin_weight} for two different emotion network architectures; ResNet-50 \cite{He2016_ResNet} and SWIN-B \cite{Liu2021_SwinTransformer}.

\begin{figure*}[t]
    \offinterlineskip
    \centering
    \includegraphics[width=0.18\columnwidth]{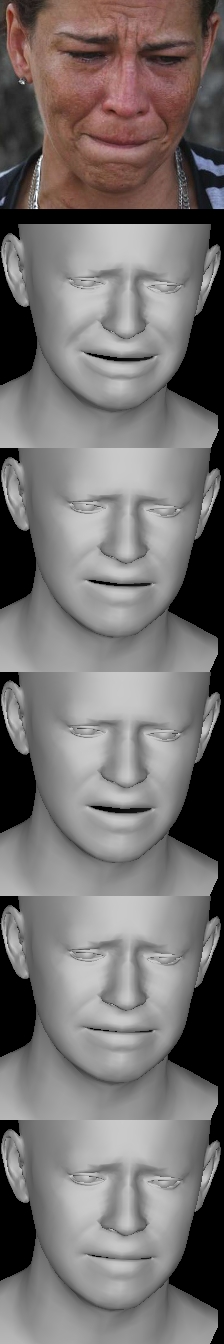}
    \includegraphics[width=0.18\columnwidth]{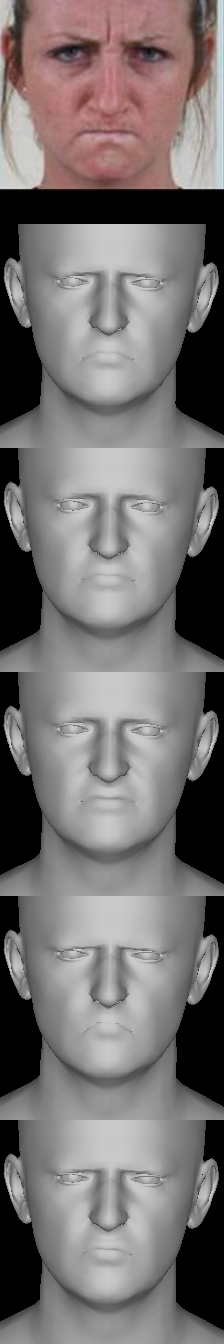} 
    \includegraphics[width=0.18\columnwidth]{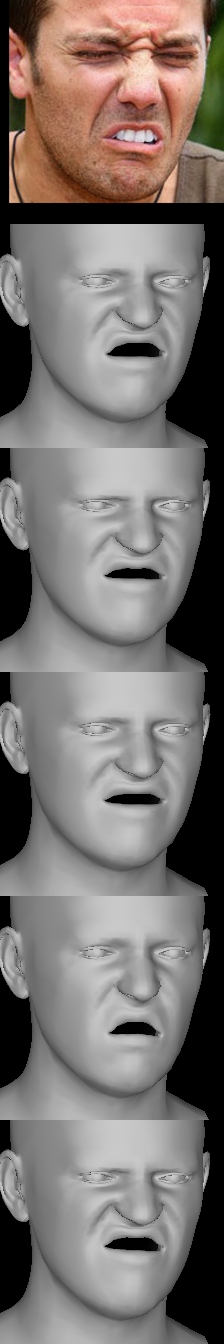} 
    \includegraphics[width=0.18\columnwidth]{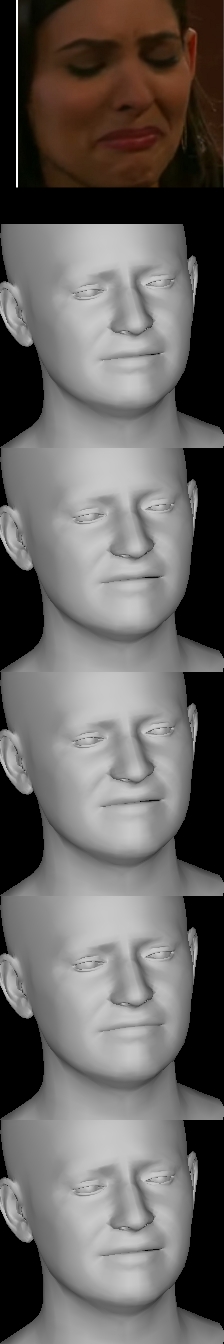} 
    \includegraphics[width=0.18\columnwidth]{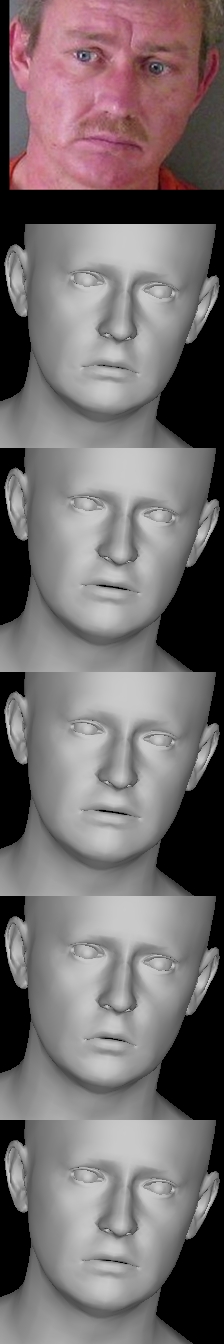} 
    \includegraphics[width=0.18\columnwidth]{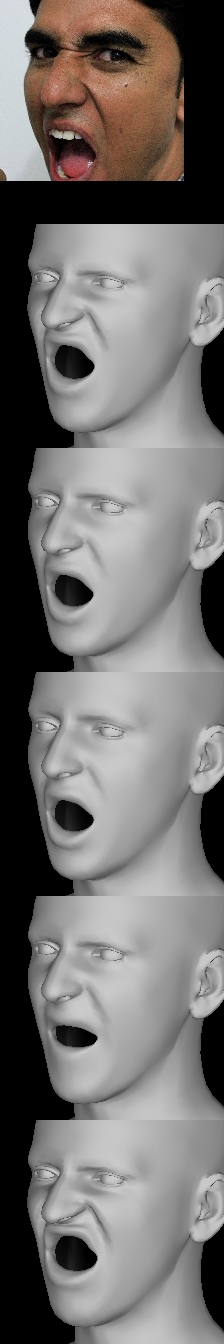} 
    \includegraphics[width=0.18\columnwidth]{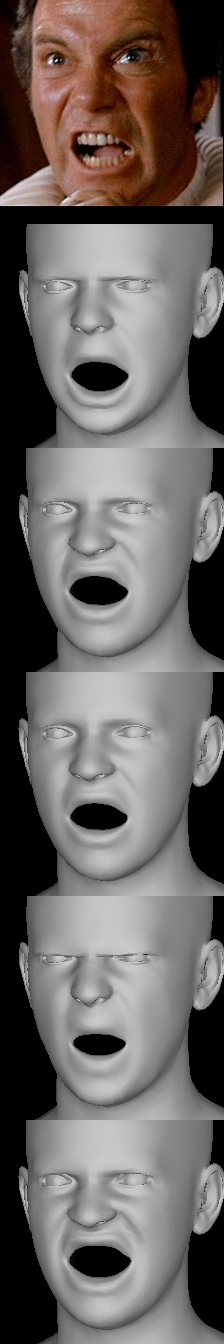} 
    \includegraphics[width=0.18\columnwidth]{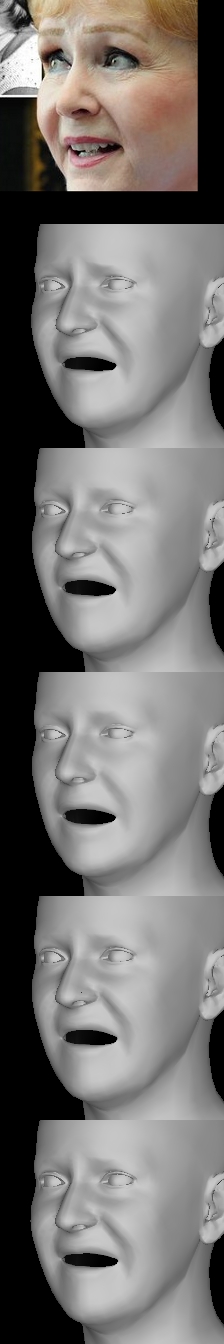} 
    \includegraphics[width=0.18\columnwidth]{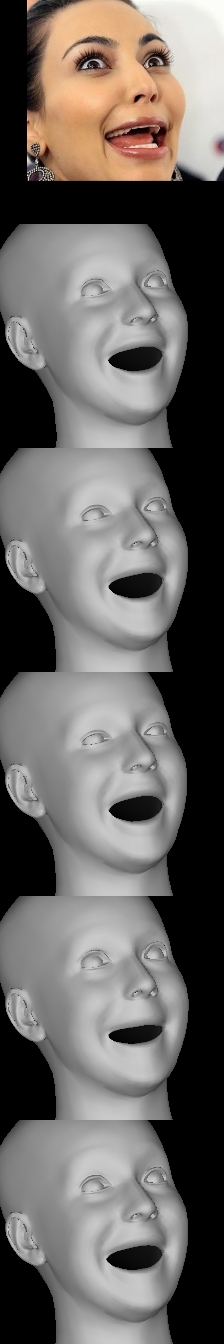} 
    \includegraphics[width=0.18\columnwidth]{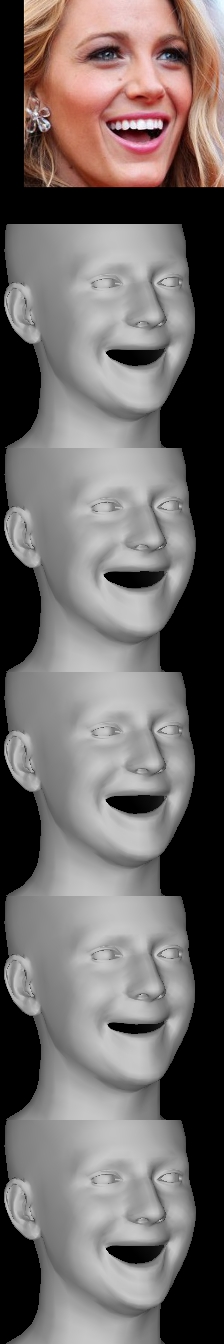} 
    \includegraphics[width=0.18\columnwidth]{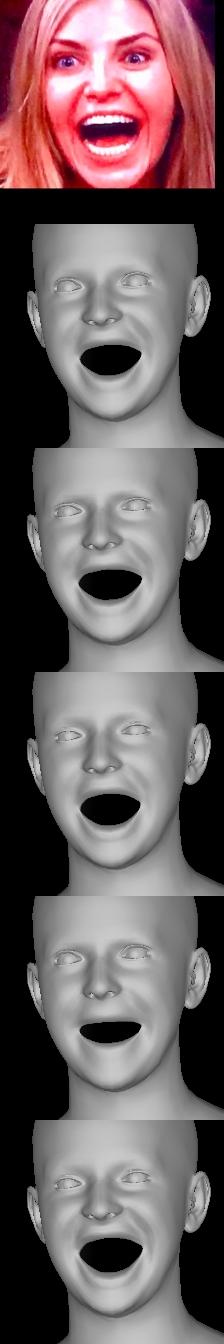} 

    \caption{A visual comparison of model with different changes. First row consists of input images. The next three rows use different metrics for evaluating emotion similarity - L2 (\model), L1 and cosine similarity. As you can observe, the selection of the metric is not critical for performance. The following row drops the relative landmark losses (mouth closure, eye closure and lip corner distance). Observe that this has a negative effect on the samples, particularly the mouth region. Final row is \model model trained on the same data as DECA instead of AffectNet. You can see that it achieves a very similar result compared to \model trained on AffectNet. This highlights an interesting finding - once an emotion recognition network has been trained, it can be used for supervision even on datasets that do not strictly guarantee a balanced representation of emotional states, such as face recognition datasets. 
    }
    \label{fig:additional}
\end{figure*}

\qheading{Additional ablations:}
We further evaluate the impact of the similarity metric used for the emotion similarity, the effect of adding a landmark reprojection error to the loss function, and the effect of the relative landmark losses (mouth closure, lip corner distance and eye closure).
Finally, we analyze the effect of using DECA's training data instead of AffectNet. 
You can see the results in \figref{fig:additional}.

\section{Emotional retargeting}
\begin{figure*}[t]
    \offinterlineskip
    \centering
    
    \includegraphics[width=0.25\columnwidth]{fig/retargetting/white.jpg}
    \includegraphics[width=0.25\columnwidth]{fig/retargetting/white.jpg} 
    \includegraphics[width=0.25\columnwidth]{fig/retargetting/white.jpg} 
    \includegraphics[width=0.25\columnwidth]{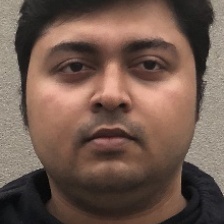} 
    \includegraphics[width=0.25\columnwidth]{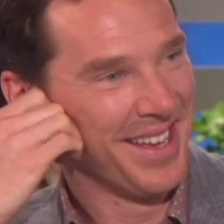} 
    \includegraphics[width=0.25\columnwidth]{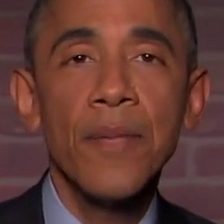} \\ 
    
    \includegraphics[width=0.25\columnwidth]{fig/retargetting/white.jpg}
    \includegraphics[width=0.25\columnwidth]{fig/retargetting/white.jpg} 
    \includegraphics[width=0.25\columnwidth]{fig/retargetting/white.jpg} 
    \includegraphics[width=0.25\columnwidth]{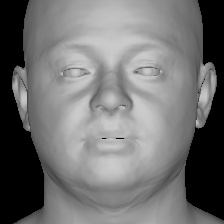}
    \includegraphics[width=0.25\columnwidth]{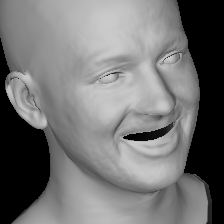} 
    \includegraphics[width=0.25\columnwidth]{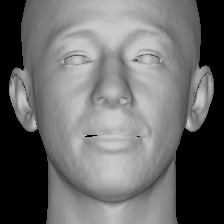} \\

    \includegraphics[width=0.25\columnwidth]{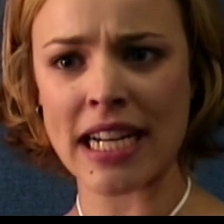} 
    \includegraphics[width=0.25\columnwidth]{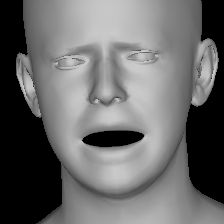}
    \includegraphics[width=0.25\columnwidth]{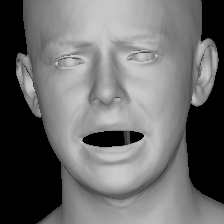} 
    \includegraphics[width=0.25\columnwidth]{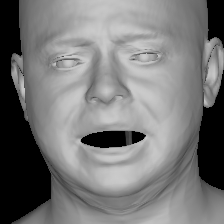} 
    \includegraphics[width=0.25\columnwidth]{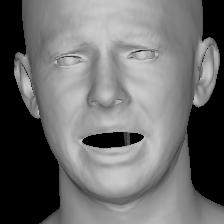} 
    \includegraphics[width=0.25\columnwidth]{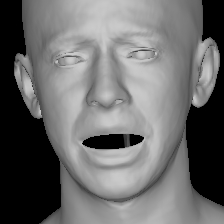} \\
    
    \includegraphics[width=0.25\columnwidth]{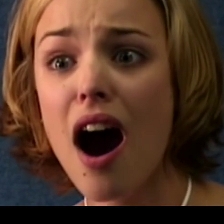} 
    \includegraphics[width=0.25\columnwidth]{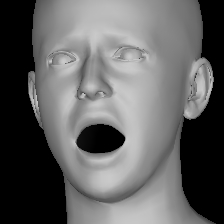}
    \includegraphics[width=0.25\columnwidth]{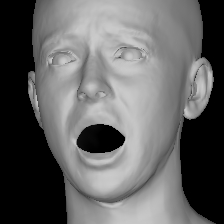} 
    \includegraphics[width=0.25\columnwidth]{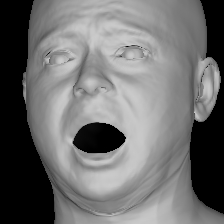} 
    \includegraphics[width=0.25\columnwidth]{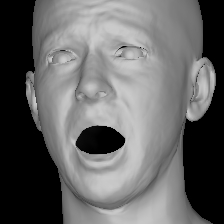} 
    \includegraphics[width=0.25\columnwidth]{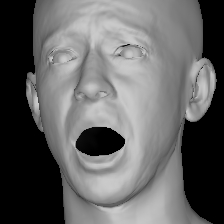} \\
    
    \includegraphics[width=0.25\columnwidth]{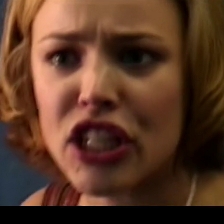} 
    \includegraphics[width=0.25\columnwidth]{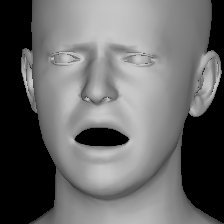}
    \includegraphics[width=0.25\columnwidth]{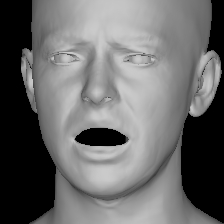} 
    \includegraphics[width=0.25\columnwidth]{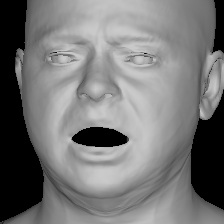} 
    \includegraphics[width=0.25\columnwidth]{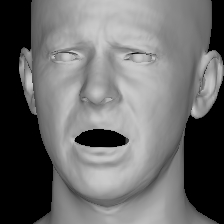} 
    \includegraphics[width=0.25\columnwidth]{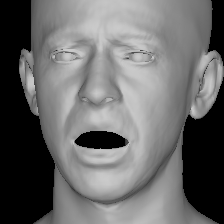} \\
    
    \includegraphics[width=0.25\columnwidth]{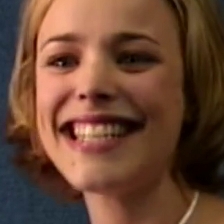} 
    \includegraphics[width=0.25\columnwidth]{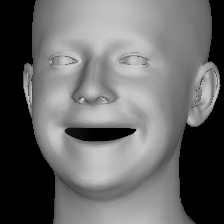}
    \includegraphics[width=0.25\columnwidth]{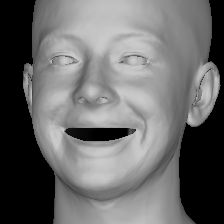} 
    \includegraphics[width=0.25\columnwidth]{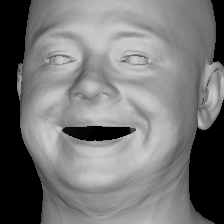} 
    \includegraphics[width=0.25\columnwidth]{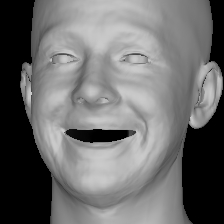} 
    \includegraphics[width=0.25\columnwidth]{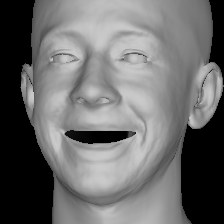} \\
    
    \includegraphics[width=0.25\columnwidth]{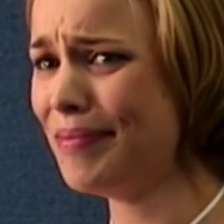} 
    \includegraphics[width=0.25\columnwidth]{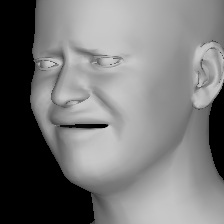}
    \includegraphics[width=0.25\columnwidth]{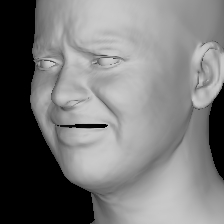} 
    \includegraphics[width=0.25\columnwidth]{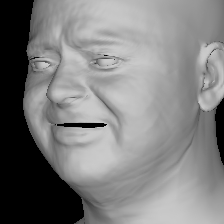} 
    \includegraphics[width=0.25\columnwidth]{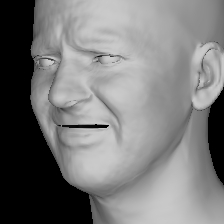} 
    \includegraphics[width=0.25\columnwidth]{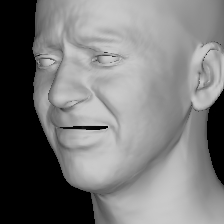} \\
    
    \includegraphics[width=0.25\columnwidth]{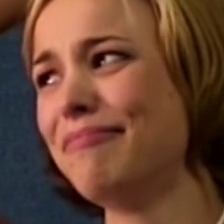} 
    \includegraphics[width=0.25\columnwidth]{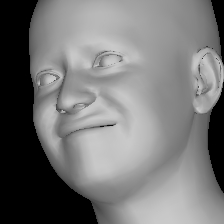}
    \includegraphics[width=0.25\columnwidth]{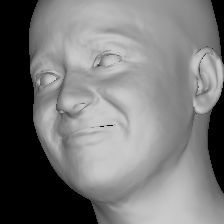} 
    \includegraphics[width=0.25\columnwidth]{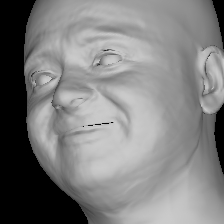} 
    \includegraphics[width=0.25\columnwidth]{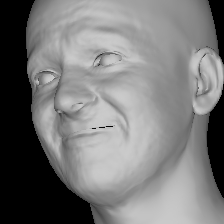} 
    \includegraphics[width=0.25\columnwidth]{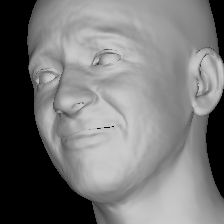} \\
    \includegraphics[width=0.25\columnwidth]{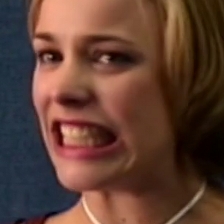} 
    \includegraphics[width=0.25\columnwidth]{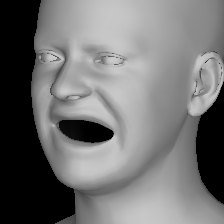}
    \includegraphics[width=0.25\columnwidth]{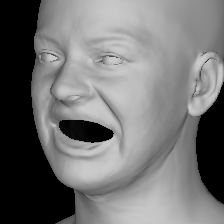} 
    \includegraphics[width=0.25\columnwidth]{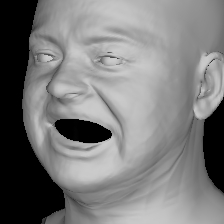} 
    \includegraphics[width=0.25\columnwidth]{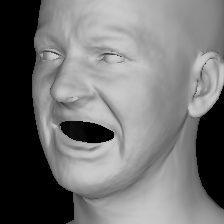} 
    \includegraphics[width=0.25\columnwidth]{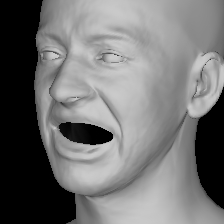} \\

    \includegraphics[width=0.25\columnwidth]{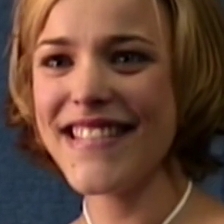} 
    \includegraphics[width=0.25\columnwidth]{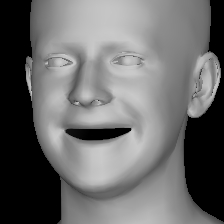}
    \includegraphics[width=0.25\columnwidth]{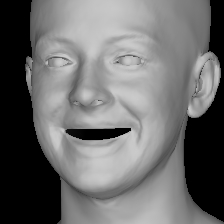} 
    \includegraphics[width=0.25\columnwidth]{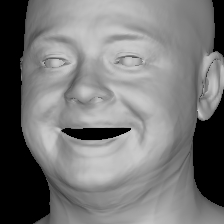} 
    \includegraphics[width=0.25\columnwidth]{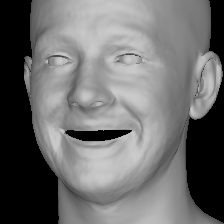} 
    \includegraphics[width=0.25\columnwidth]{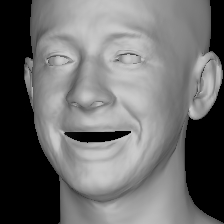} \\
    
    \caption{\textbf{Emotional retargetting.} From left to right. The input image, coarse reconstruction, detailed reconstruction, emotion retargeted to the coarse identity above. Observe that while the identity and the person-specific detailed displacements change with the source actor, the emotion fidelity is preserved. For the entire sequence in motion, please see the supplementary video.
    }
    \label{fig:retargeting}
\end{figure*}

\model regresses FLAME \cite{FLAME:SiggraphAsia2017} parameters and expression dependent geometric details.
The disentanglement of the coarse identity and expression geometry and the identity and expression dependent details allows us to animate \model's reconstructions. 
We demonstrate this by animating a source 3D face using a video sequence of another actor.
Figure~\ref{fig:retargeting} demonstrates two things, first, \model reconstructions convey emotions of the source images, and second, the animated faces of other subjects convey the same emotion. 
The emotional fidelity hence is preserved in the animated face of the other subject.

\section{Emotion retrieval}
\label{sec:retrieval}
Our work relies on the following key hypothesis. 
The emotion recognition networks learn a useful embedding of emotion. 
The following properties are desirable: 
\begin{itemize}
    \item Images of faces with similar expressions conveying similar emotions are close in this embedding space.
    \item Images of faces with dissimilar expressions/emotions are farther apart in this space.
    \item Invariance to pose, identity and lighting and background.
\end{itemize}

\begin{figure*}[t]
    \offinterlineskip
    \centering
    \includegraphics[width=2.1\columnwidth]{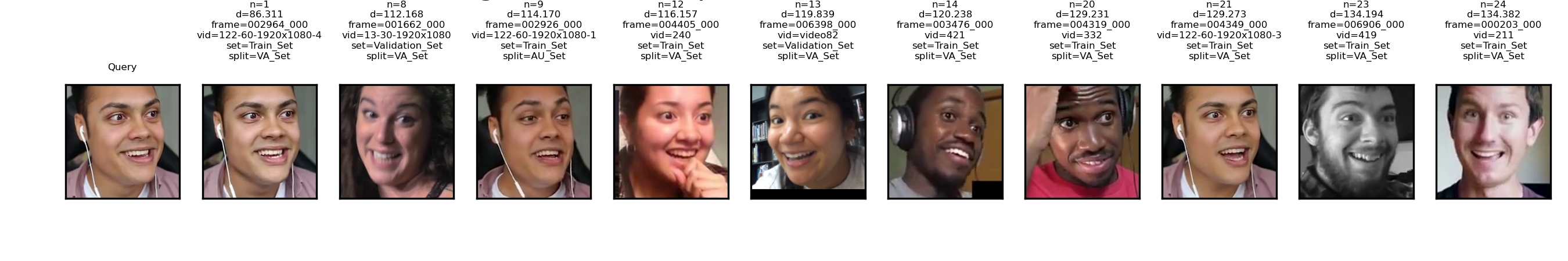} \\
    \includegraphics[width=2.1\columnwidth]{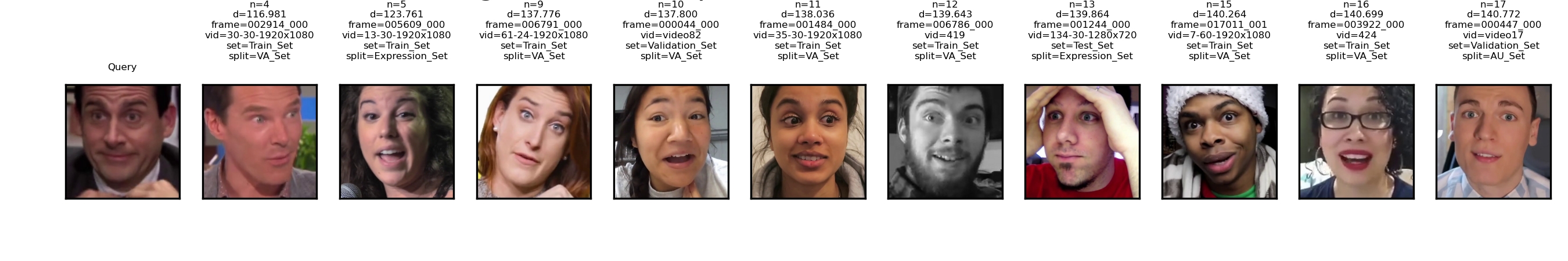} \\
    \includegraphics[width=2.1\columnwidth]{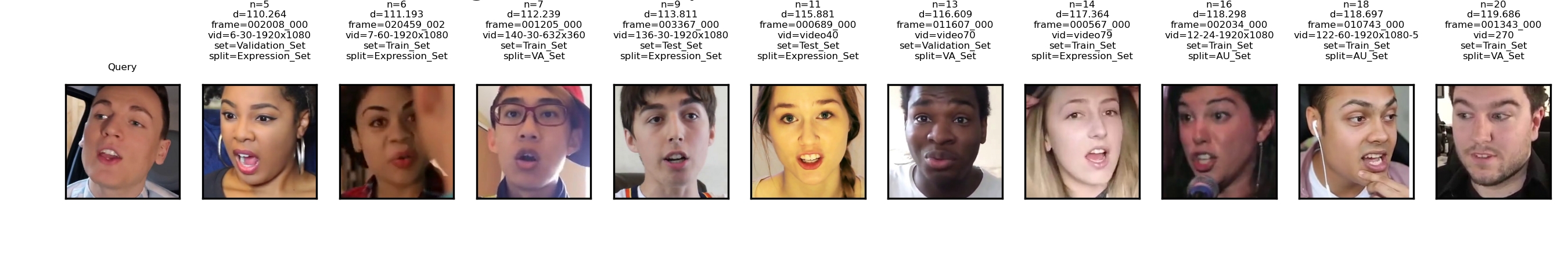} \\
    \includegraphics[width=2.1\columnwidth]{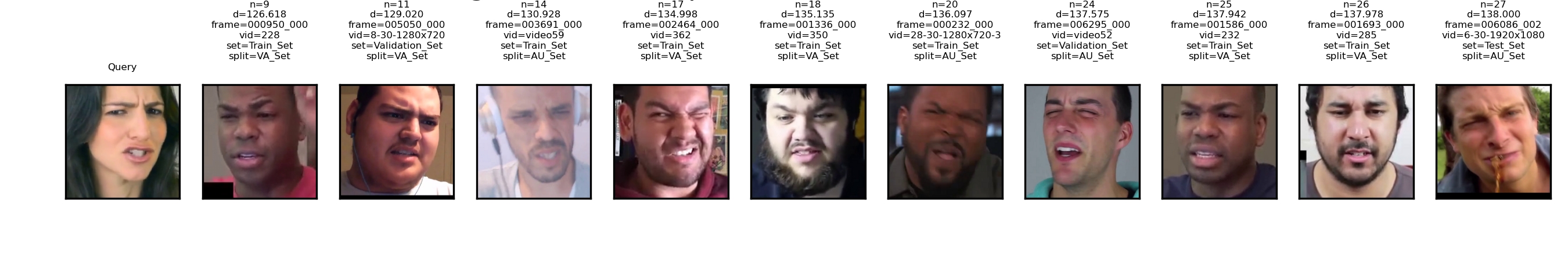} \\
    \includegraphics[width=2.1\columnwidth]{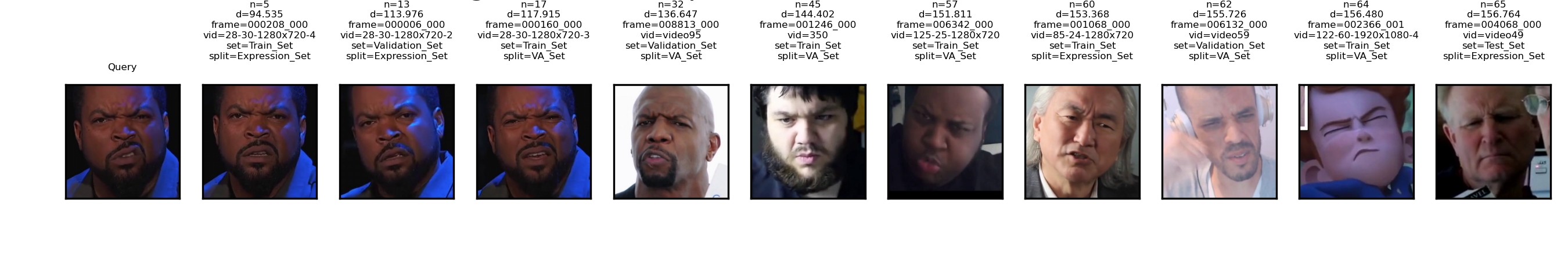} \\
    \includegraphics[width=2.1\columnwidth]{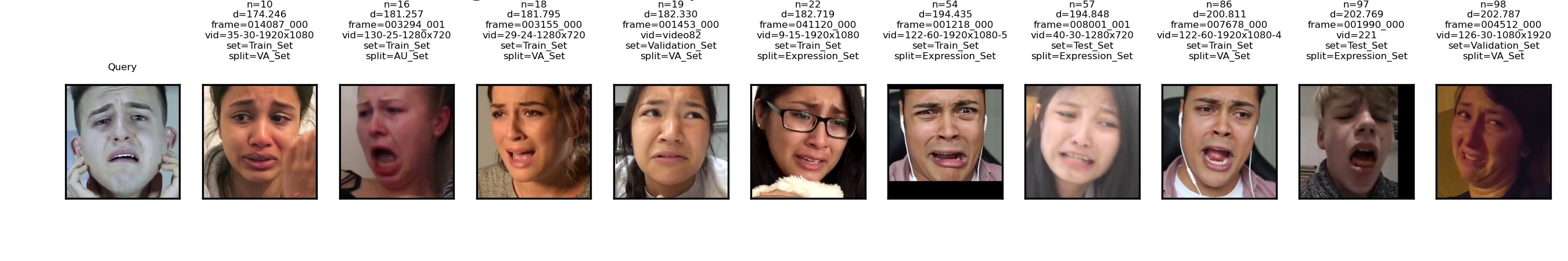} \\
    \includegraphics[width=2.1\columnwidth]{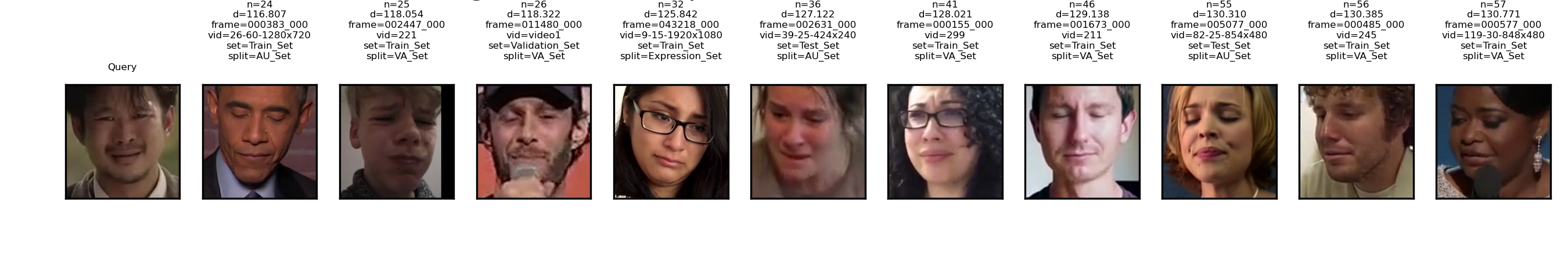} \\
    \caption{Examples of nearest neighbor retrieval using the EmoNet \cite{toisoul2021estimation} feature. We searched for up to 100 neighbors. We ony include up to 1 NN per video to avoid retrieving consecutive frames. Left: query image, Right: ordered nearest neighbors from different clips. Observe how all of the retrieved faces communicate very similar emotional content.
    }
    \label{fig:retrieval_emonet_feat}
\end{figure*}

\begin{figure*}[t]
    \offinterlineskip
    \centering
    \includegraphics[width=2.1\columnwidth]{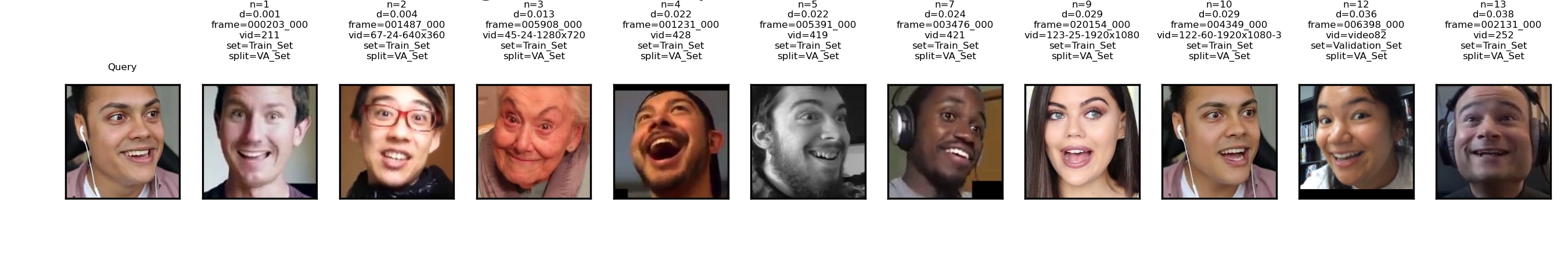} \\
    \includegraphics[width=2.1\columnwidth]{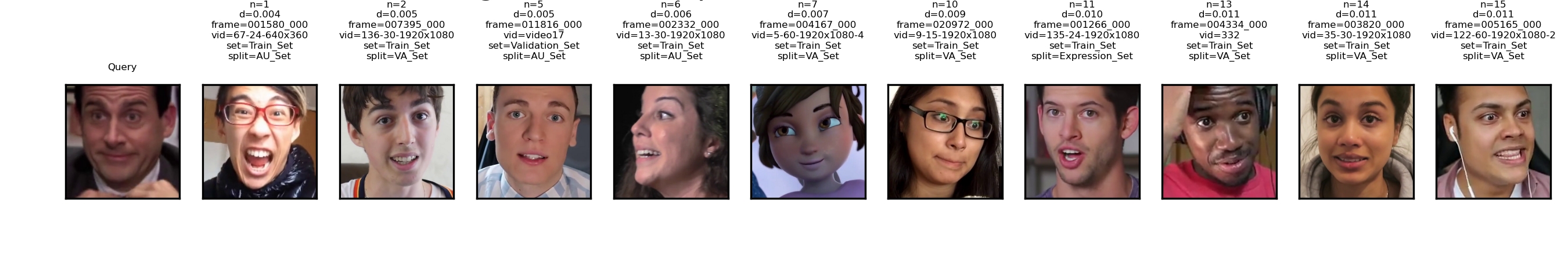} \\
    \includegraphics[width=2.1\columnwidth]{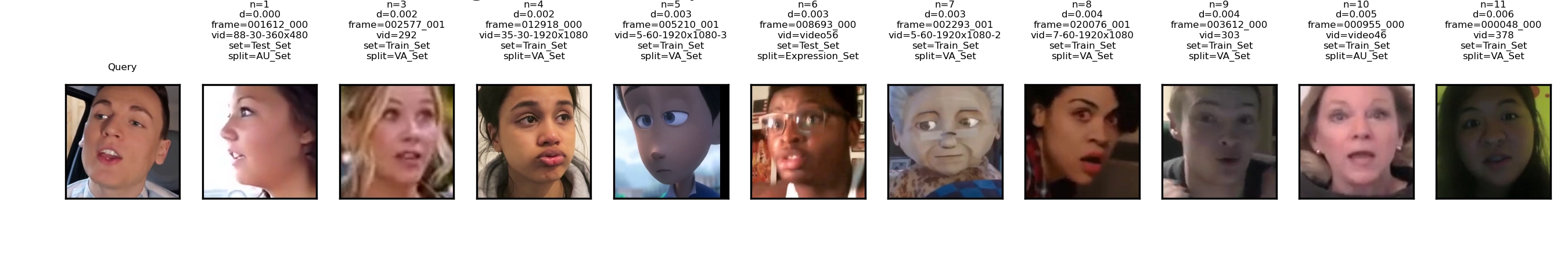} \\
    \includegraphics[width=2.1\columnwidth]{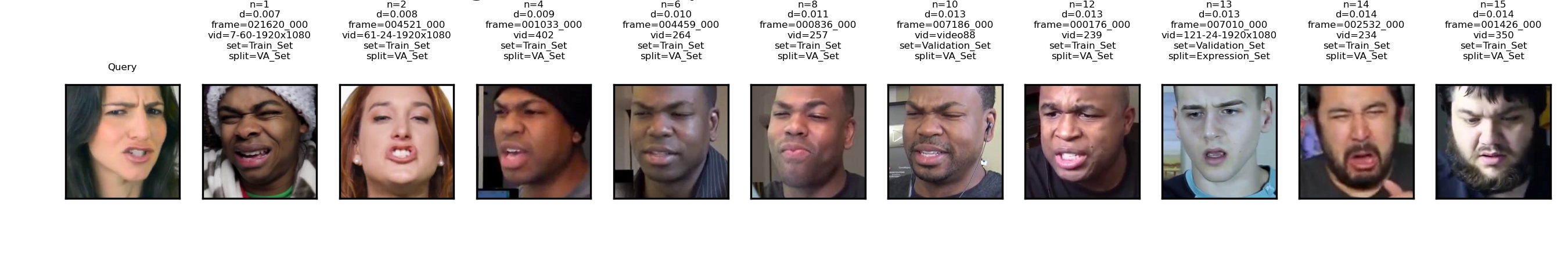} \\
    \includegraphics[width=2.1\columnwidth]{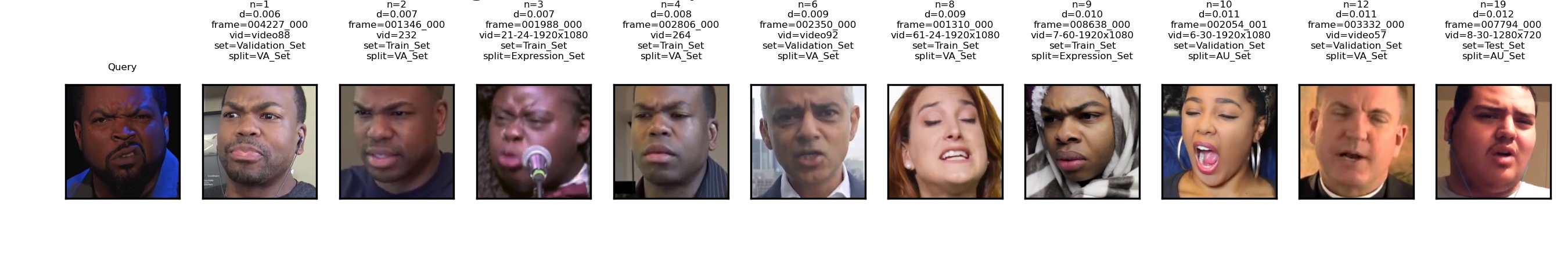} \\
    \includegraphics[width=2.1\columnwidth]{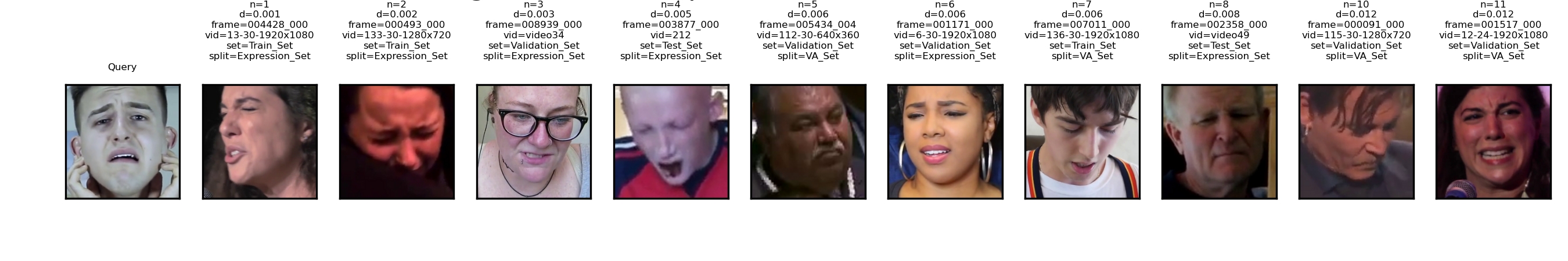} \\
    \includegraphics[width=2.1\columnwidth]{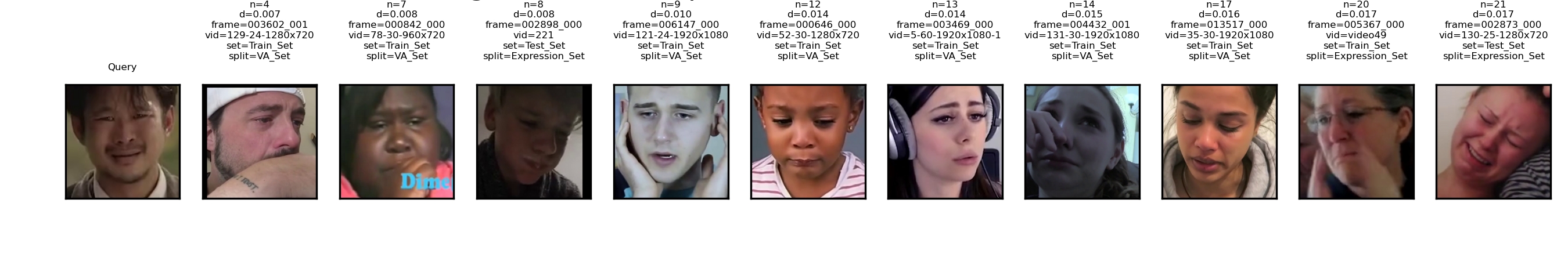} \\
    \caption{Examples of nearest neighbor retrieval using the ground truth annotated valence and arousal space on the AffWild2 \cite{Aff-Wild2} dataset. While the retrieved faces do have some degree of similarity, the quality of retrieval compared to the EmoNet feature is lower.
    }
    \label{fig:retrieval_va}
\end{figure*}

We employ the publicly released model of EmoNet \cite{toisoul2021estimation} and use the 256-dimensional feature output of the last convolutional layer as emotion embedding. 
We then extract the emotion embedding for faces in the Aff-Wild2 video dataset \cite{Aff-Wild2}. 
For the emotion retrieval given an image, we seek the nearest neighbors w.r.t. L2 distance metric in the dataset.
Figure~\ref{fig:retrieval_emonet_feat} shows the 10 nearest neighbors for multiple images. 
For comparison, we repeat the process for the ground truth (GT) valence and arousal labels of the Aff-Wild2 dataset in \figref{fig:retrieval_va}.

\end{document}